\documentclass{article} % For LaTeX2e
\usepackage{iclr2025_conference,times}

\usepackage{amsmath,amsfonts,bm}

\def\eqref#1{equation~\ref{#1}}
\def\1{\bm{1}}

\DeclareMathAlphabet{\mathsfit}{\encodingdefault}{\sfdefault}{m}{sl}
\SetMathAlphabet{\mathsfit}{bold}{\encodingdefault}{\sfdefault}{bx}{n}

\usepackage{hyperref}
\usepackage{url}

\usepackage{inconsolata}
\usepackage{url}
\usepackage{xltabular}
\usepackage{xtab,booktabs}
\usepackage{tabularx}
\usepackage{tabularray}
\usepackage{graphicx}
\usepackage{subcaption}
\usepackage{caption}
\usepackage{amsmath}
\usepackage{amssymb}
\usepackage{mathtools}
\usepackage{amsthm}
\usepackage{multirow}
\usepackage{makecell}
\usepackage{graphicx}
\usepackage{enumitem}
\usepackage{wrapfig}
\usepackage{xspace}
\title{Towards World Simulator: Crafting Physical Commonsense-Based Benchmark for Video Generation}

\iclrfinalcopy

\author{Fanqing Meng$^{*, 1, 2}$, Jiaqi Liao$^{*, 2}$, Xinyu Tan,  \textbf{Wenqi Shao}$^{2, \dagger}$, Quanfeng Lu$^{2}$, \textbf{Kaipeng Zhang}$^{2}$ \\
  \textbf{Yu Cheng}$^{4}$,   \textbf{Dianqi Li},  \textbf{Yu Qiao}$^{2}$, \textbf{Ping Luo}$^{3, 2, \dagger}$ \\\\
$^{1}$Shanghai Jiao Tong University \quad  $^{2}$OpenGVLab, Shanghai AI Laboratory \\
$^{3}$The University of Hong Kong  \quad $^{4}$ The Chinese University of Hong Kong \\
}

\newcommand{\bench}{\textsl{PhyGenBench}\xspace}
\newcommand{\eval}{\textsl{PhyGenEval}\xspace}

\begin{document}

\maketitle

\begin{center}
\vspace{-0.4in}
\textcolor{brown}{Project Page:}\,\,\href{https://phygenbench123.github.io/}{\color{brown}{https://phygenbench123.github.io/}}
\end{center}

\renewcommand{\thefootnote}{\fnsymbol{footnote}}
{\let\thefootnote\relax\footnotetext{
\noindent \hspace{-5mm}$\dagger$ Corresponding Authors: shaowenqi@pjlab.org.cn; pluo@cs.hku.hk \\
\noindent \hspace{-5mm}\quad \quad $*$ Equal contribution \\
}   }

\begin{abstract}

Text-to-video (T2V) models like Sora have made significant strides in visualizing complex prompts, which is increasingly viewed as a promising path towards constructing the universal world simulator. Cognitive psychologists believe that the foundation for achieving this goal is the ability to understand intuitive physics. However, the capacity of these models to accurately represent intuitive physics remains largely unexplored. To bridge this gap, we introduce \bench, a comprehensive \textbf{Phy}sics \textbf{Gen}eration \textbf{Bench}mark designed to evaluate physical commonsense correctness in T2V generation. \bench comprises 160 carefully crafted prompts across 27 distinct physical laws, spanning four fundamental domains, which could comprehensively assesses models' understanding of physical commonsense. Alongside \bench, we propose a novel evaluation framework called \eval. This framework employs a hierarchical evaluation structure utilizing appropriate advanced vision-language models and large language models to assess physical commonsense. Through \bench and \eval, we can conduct large-scale automated assessments of T2V models' understanding of physical commonsense, which align closely with human feedback. Our evaluation results and in-depth analysis demonstrate that current models struggle to generate videos that comply with physical commonsense. Moreover, simply scaling up models or employing prompt engineering techniques is insufficient to fully address the challenges presented by \bench (e.g., dynamic physical phenomenons). We hope this study will inspire the community to prioritize the learning of physical commonsense in these models beyond entertainment applications. We release the data and codes at \url{https://github.com/OpenGVLab/PhyGenBench}

% but their ability to capture intuitive physics remains a critical challenge in their development as universal world simulators. To address this gap, we introduce \bench, a comprehensive benchmark designed to evaluate physical commonsense correctness in T2V generation. \bench comprises $160$ carefully crafted prompts across $27$ distinct physical laws, spanning four fundamental domains: optics, mechanics, thermodynamics, and material properties. Along with \bench, we propose \eval, a novel hierarchical evaluation framework that leverages advanced vision-language models and GPT-4o to assess physical commonsense correctness in generated content. 

% Notably, \eval exhibits a high correlation with human feedback, outperforming existing approaches in evaluating physical plausibility. We conduct extensive evaluations of popular T2V models using \bench, providing in-depth analyses of their performance. This work not only establishes a robust methodology for assessing physical commonsense correctness in T2V models but also aims to inspire the AI community to prioritize the learning of physical knowledge in these models beyond entertainment applications. 

% Our contributions include: (1) we propose the \bench, a benchmark for evaluating T2V models' adherence to physical commonsense. (2) we introduce the \eval, an effective framework for assessing physical alignment in generated videos. (3) we conduct comprehensive evaluation and analysis on \bench, offering insights to guide future research in this field.

\end{abstract}

\begin{figure*}[htbp]
  \centering
  \scalebox{0.99}{
  \includegraphics[width=\linewidth]{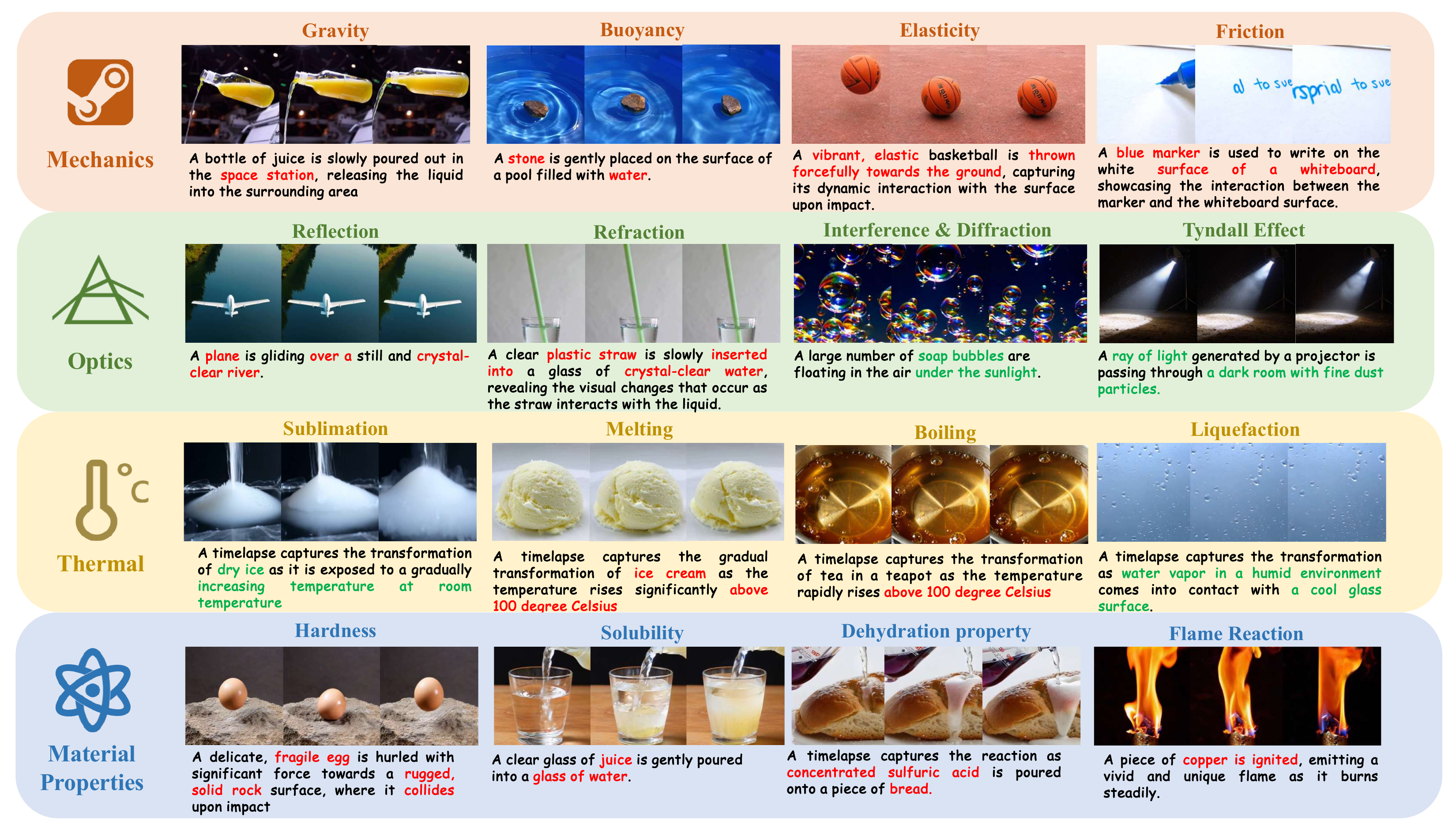}
  }
  % \vspace{-40pt}
  \caption{Samples of videos generated by Kling or Gen-3 in \bench with $4$ different aspects. The results show that current T2V models struggle to generate videos that align with physical commonsense (e.g., the lack of a plane's reflection in water in the first video of the second row). }
  \label{fig:demo}
  \vspace{-10pt}
\end{figure*}

\section{Introduction}

% Text-to-video (T2V) models like SORA have significantly advanced the ability to visualize complex ideas and scenes based on textual input \citep{yang2024cogvideox, wang2023lavie}. However, beyond simply creating videos from text, it is vital for these models to align with physical principles. This alignment ensures the models can generate accurate object interactions or realistic lighting in scenes, which is fundamental for developing realistic simulations of the real world \citep{Milliere2024VideoGenerationWorldSimulators}. However, even the most advanced models, such as Kling \citep{}, struggle to accurately generate some extremely simple physical scenarios, as illustrated in figure \ref{fig:demo} .
Text-to-video (T2V) models such as Sora have made significant strides in visualizing complex ideas and scenes based on textual input \citep{yang2024cogvideox, wang2023lavie}. These advancements are increasingly viewed as a promising path towards constructing universal simulators of the physical world, which holds immense promise for video generation \citep{zhu2024sora}, autonomous driving \citep{gao2024vista}, and the development of embodied agents \citep{mazzaglia2024multimodal}.
% \dianqi{should mention the application of world simulator (and citations)}
% Cognitive psychology posits that achieving this goal first requires intuitive physics \citep{margoni2024violation}, emphasizing the visual and interactive plausibility of rendered scenes rather than strict physical accuracy. Notably, even human infants demonstrate correct intuitive physical cognition \citep{wood2024object,battaglia2013simulation}. 
Cognitive psychology posits that intuitive physics, which is demonstrated even by human infants  \citep{wood2024object,battaglia2013simulation}, is essential for achieving this goal. Intuitive physics emphasizes rendered scenes should be visually and interactively natural to humans, rather than adhere to strict physical accuracy.
% The intuitive physics emphasizes the visual and interactive plausibility of rendered scenes rather than strict physical accuracy \citep{margoni2024violation}.
Consequently, on the path towards developing a world simulator \citep{xiang2024pandora},video generation should first be capable of accurately reproducing simple yet fundamental physical phenomenons.
% \dianqi{logic is a little bit verbose and tangled: "Cognitive psychology" \textrightarrow "Notable" \textrightarrow "Consequently" \textrightarrow "However". Can consolidate the descriptions before "However". }
However, even state-of-the-art models trained on vast resources \citep{tan2024vidgen} encounter difficulties in correctly generating seemingly trivial physical phenomenons, as depicted in Figure \ref{fig:demo}, the model fails to understand that the stone should sink in water. This clear pitfall shows a substantial gap between current video generation models' and human's understanding of basic physics. It reveals how far these models are from being true world simulators.

Given this context, it becomes important to assess the extent to which current T2V models can capture intuitive physics in their generated outputs. This requires the development of comprehensive evaluation frameworks that beyond traditional metrics. While numerous Text-to-Video (T2V) evaluation benchmarks have emerged \citep{sun2024t2v,huang2024vbench}, they primarily focus on various qualities of generated videos (e.g., motion smoothness, background consistency) or spatial relationships, failing to address the critical issue of whether the generated videos adhere to fundamental physical laws. 
%While numerous Text-to-Video (T2V) evaluation benchmarks have emerged \citep{sun2024t2v,huang2024vbench,liao2024evaluation}, they primarily focus on various qualities of generated videos (e.g., motion smoothness, background consistency)
% % \dianqi{"video motion quality" and "dynamic aspects" terms are too vague, they should also be part of physical principles. The point we should say is that they can't evaluate physics} or spatial relationships, failing to address the critical issue of whether the generated videos adhere to fundamental physical laws. 
Although some studies have explored the alignment of generated videos with dynamic motions naturalness \citep{bansal2024videophy}, their benchmarks fail to succinctly capture fundamental physical laws or propose sufficiently robust evaluation methods.
% \dianqi{if that's the case, then we shouldn't say they evaluated "physical laws" which would undermine our contribution. You can change it with something like "dynamic motions".}. 
Therefore, the development of benchmarks and evaluation methodologies specifically tailored to assess intuitive physics in generated videos remains a critical yet largely unexplored frontier.

There are two challenges impeding the evaluation of physical commonsense in T2V models. On one hand, there is a lack of benchmarks focused on evaluating physical commonsense. This requires selecting semantically simple physical phenomenons that exhibit clear physical phenomena, allowing for accurate assessment by either humans or machines. On the other hand, there is a lack of corresponding evaluation metrics. Traditional metrics like FVD \citep{unterthiner2018towards} exhibit limitations in detecting implausible motions \citep{brooks2022generating} and necessitate reference videos, which are often challenging to procure for novel scenes. Recent studies have used video-based VLMs for comprehensive video evaluation \citep{he2024mantisscore,sun2024t2v}. However, they often struggle to correctly assess physical commonsense. This limitation stems from their inadequate understanding of physical laws \citep{jassim2023grasp} and the fact that these methods are not specifically designed to evaluate physical laws.

% while Vision-Language Model (VLM)-based evaluation approaches \citep{he2024mantisscore} demonstrate limited generalizability and specific in assessing physical commonsense accuracy. \dianqi{any references for the last statement? Otherwise it feels a little bit overclaimed, as VLM evaluation via prompt basic is targeting generalizability. In addition, our approaches also utilized VLM for evaluation, right? (then this statement is not suitable)}

% \dianqi{There are two challenges, missing benchmark and evaluation method. You don't need paragraphs to refer to them, you can just compose them into the same paragraph.}
To address these challenges, we propose \bench and \eval to automate the evaluation of physical commonsense understanding capability from T2V models. \bench is designed to evaluate physical commonsense based on fundamental physical laws in text-to-video generation. Inspired by~\citep{halliday2013fundamentals}, we categorize physical commonsense in the world into four main areas: mechanics, optics, thermal, and material properties. And we identify significant physical laws and easily observable physical phenomenons for each category, resulting in comprehensive $27$ physical laws and $160$ validated prompts in the proposed benchmark. 
% \dianqi{I reorg the statement orders: first important highlight \textrightarrow then some details. In addition, why mention "27", "160" valided prompts? what do you want to say from the numbers?} 
% Specifically, \bench constructs a comprehensive but simple set of prompts reflecting physical commonsense, which are sufficiently clear for evaluation. 
Specifically, we start from fundamental physical laws. Through brainstorming, we construct prompts that easily reflect physical laws using sources like textbooks \citep{harjono2020interactive}. This process results in a comprehensive but simple set of prompts reflecting physical commonsense, which are sufficiently clear for evaluation.
As shown in Figure \ref{fig:demo}, the correctness of physical commonsense in \bench can be observed through clear phenomena (e.g., \textit{plane should have reflections in water}) 
% \dianqi{Then how do we construct the prompt from a high-level idea?}. \dianqi{also I changed the logic again to make it more clear} \dianqi{feel like we need to have a connection in the later methodologies section on how we bridge physical commonsense prompts with fundamental physics laws.} 
On the other hand, benefiting from the simple yet clear physical phenomena in \bench prompts, we can propose \eval, which is a novel video evaluation framework for assessing physical commonsense correctness in \bench.
% \dianqi{still missing the motivation connection from Benchmark to Evaluation, e.g., you already mentioned "prompts" in the benchmark. shouldn't those prompts be used for evaluation? or what they are used for?}, 
% \\eval is a novel video evaluation framework for assessing physical commonsense correctness in \bench. 
\eval first uses GPT-4o to analyze physical laws in text, addressing the poor understanding of physical common sense in video-based VLMs. Moreover, considering that previous evaluation metrics did not specifically target physical correctness, we propose a three-tier hierarchical evaluation strategy for this aspect, transitioning from image-based to comprehensive video analysis: single image, multiple images, and full video stages. Each stage employs distinct VLMs along with custom instructions generated by GPT-4o to form judgments. By combining \bench and \eval, we can efficiently evaluate different T2V models' understanding of physical commonsense at scale, producing results highly consistent with human feedback.

The contributions of our work are three-fold.  \textbf{i): }We proposed \bench, which compasses a wide range of clear physical phenomenons and explicit physical laws. This benchmark can comprehensively measure whether T2V models understand intuitive physics and indirectly assess their gap from world simulator capabilities
% \dianqi{did we really assess the gap from the world simulator? How do you qualify this part?}.
% making it easy to evaluate whether T2V models conform to physical commonsense
% \dianqi{contribution is too superficial? Just making things easy?}
\textbf{ii): }Along with the benchmark, we propose an automated evaluation framework - \eval, which overcomes the challenges of assessing the correctness of physical commonsense with other metrics and demonstrates high consistency with human feedback on \bench, enabling users to conduct large-scale automated testing of various T2V models.
% iii) We conduct extensive evaluations and in-depth analyses of popular T2V models using \eval and \bench. The results indicate that currently, all T2V models struggle to achieve satisfactory outcomes, and a straightforward scaling-up approach fails to address all issues. \dianqi{Can you be more specific for the insights? There are no insights from the statement, the only thing readers would know is all existing models performed poorly on our benchmark.}
\textbf{iii): }We conduct extensive evaluations of popular T2V models, even the best-performing model, Gen-3, scores only $0.51$. This indicates that current models are still far from functioning as world simulators. Based on our evaluation results, we conduct an in-depth analysis and discover that addressing issues such as dynamics is still challenging through prompt engineering or simply scaling up model.
We hope this work inspires the community to focus on the learning of physical commonsense in T2V models, rather than merely using them as tools for entertainment.

\section{Related work}

\subsection{Benchmarks for text-to-video generation}
The rapid advancement of text-to-video (T2V) generation models has necessitated various benchmarks for accurate assessment. Traditional works in video generation, such as FVD \citep{unterthiner2018towards}, rely on datasets like UCF-101 \citep{soomro2012ucf101} and Kinetics-400 \citep{kay2017kinetics}, which are limited in scope. Recent benchmarks, including VBench \citep{huang2024vbench} and EvalCrafter \citep{liu2024evalcrafter}, aim to comprehensively evaluate general video quality across multiple dimensions. In contrast, some studies focus on fine-grained evaluation of text-to-video (T2V) models from specific aspects. For instance, T2V-CompBench \citep{sun2024t2v} assesses compositional generation capabilities, while DEVIL \citep{liao2024evaluation} evaluates dynamic characteristics of generated videos. Although some research like VideoPhy \citep{bansal2024videophy} efforts address the  dynamic motions naturalness of video generation, their benchmarks fail to succinctly capture fundamental physical laws. Consequently, most existing works overlook this crucial aspect, which forms the foundation for realizing a world simulator. To address this gap, we introduce \bench, a benchmark designed to comprehensively measure T2V models' understanding of physical commonsense. 

\subsection{Evaluation metrics for text-to-video generation}
Conventional approaches to video quality assessment often employ metrics such as FVD \citep{unterthiner2018towards} and IS \citep{salimans2016improved}. However, the detection of unrealistic motions is difficult for them \citep{brooks2022generating}, and FVD requires a reference video that is hard to obtain for novel scenes, making it challenging to evaluate the correctness of physical commonsense. Recent studies have explored the use of advanced vision-language models (VLMs) as evaluators. For instance, VideoScore \citep{he2024mantisscore} leverages human feedback to train models for video quality assessment, while T2V-CompBench \citep{sun2024t2v} utilizes powerful models like LLaVA \citep{liu2024visual} to evaluate the correctness of spatial relationships. Although a few works demonstrate improved alignment with human judgments, they fall short in generalizing to assessments of physical commonsense correctness. To address this limitation, we introduce \eval, a novel method designed to evaluate physical commonsense correctness on \bench. We validate the efficacy of our approach through comprehensive human correlation studies.

\section{PhyGenBench}
Inspired by \citep{swartz1985concept}, we first define the following terms:
\textit{``Physical Commonsense:"} Basic intuitive understanding of how physical objects and actions behave in everyday life;
\textit{``Physical Laws:"} Universal scientific principles that describe consistent behaviors in nature;
\textit{``Physical Phenomenon:"} Observable events or processes caused by the interaction of physical laws.
The purpose of \bench is to evaluate whether T2V models understand physical commonsense, while each prompt in \bench presents a clear physical phenomenon and an underlying physical law.
% Each prompt in \bench

% In the construction, we manually select physical laws from textbooks \citep{harjono2020interactive}, with each physical law corresponding to multiple commonsense. Each commonsense is represented by a physical phenomenon.
% We first provide overall information about \bench \dianqi{placeholder} and then detail the construction process of the dataset. For more detailed information about the dataset, please refer to Section XX in the appendix.
\begin{figure*}[t!]
  \centering
  \scalebox{0.7}{
  \includegraphics[width=\linewidth]{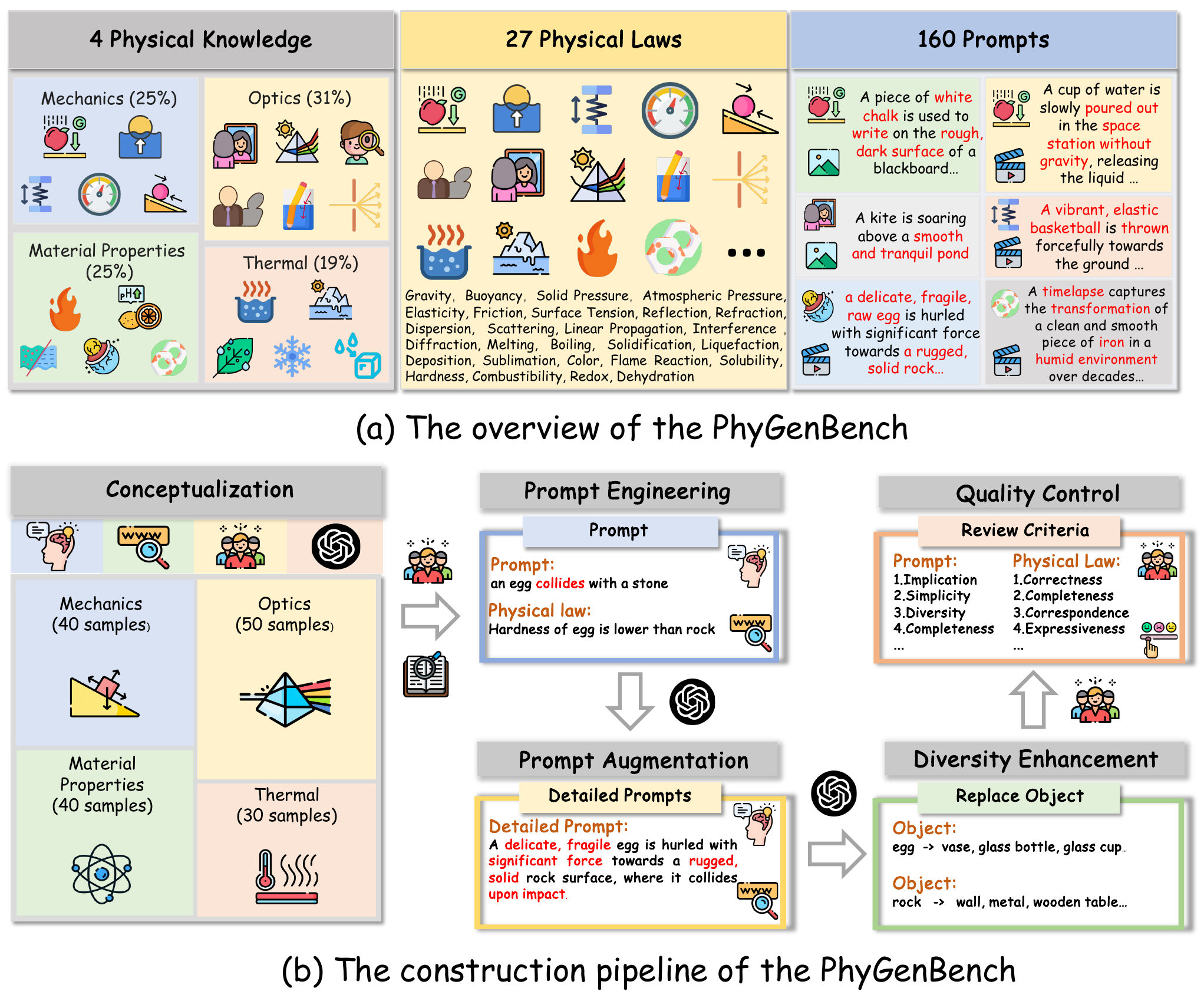}
  }
  % \vspace{-70pt}
  \caption{(a) is the overview of the proposed \bench. (b) is the \bench data pipeline, which covers four physics categories. We select key physical laws and manually craft initial prompts that reflect the corresponding physical phenomena. GPT-4o adds details and enhances diversity by varying objects. After manual review, we obtain 160 T2V prompts. }
  \label{fig:phybench}
  \vspace{-10pt}
\end{figure*}
\paragraph{Overview.} As illustrated in Figure \ref{fig:phybench} (a), \bench encompasses four major categories of physical commonsense: \textit{``Mechanics"}, \textit{``Optics"}, \textit{``Thermal"}, and \textit{``Material Properties"}. It incorporates $27$ physical phenomena with intrinsic physical laws reflected by the corresponding designed $160$ prompts: 

1. \textit{``Mechanics"} covers $7$ common mechanical laws: gravity, buoyancy, solid pressure, atmospheric pressure, elasticity, friction, and surface tension, with $40$ validated prompts. For example, we use \textsl{``A piece of iron is gently placed on the surface of the water in a tank filled with water"} to test T2V model's understanding of Buoyancy,
% \dianqi{this example is really bad, friction is not directly related to marks. Change an easily understandable one},
where the iron should sink due to its higher density compared to water.

2. \textit{``Optics"} categorizes $6$ aspects based on light phenomena: reflection, refraction, scattering, dispersion, interference $\&$ diffraction, and straight-line propagation, yielding $50$ prompts. A prompt like \textsl{``a kite soaring above a smooth and tranquil pond"} is used to test reflection generation capability.

3. \textit{``Thermal"} considers $6$ phase transitions: Solidification, Melting, Liquefaction, Boiling, deposition, Sublimation, comprising $30$ prompts. 
% \dianqi{why there is no detailed categories information?}. 
Inspired by ChronoMagicBench \citep{yuan2024chronomagic}, the vaporization (boiling) process is evaluated by the prompt \textsl{``a timelapse capturing the transformation of water as the temperature rapidly rises above $100^{\circ}C$"}.
    
4. \textit{``Material Properties"} includes $5$ physical properties (color, hardness, solubility, combustibility, and flame reaction) and $3$ chemical properties (acidity, redox potential, and dehydrating properties), resulting in $40$ prompts. We reflect material properties, e.g., \textit{``hardness"}, through the prompts with expected phenomena, e.g., \textsl{``an egg being hurled with significant force towards a rock"}, where the egg should break while the rock remains intact.

% We emphasize that while any given prompt may reflect multiple physical laws, this complicates the evaluation of physical common sense in video generation, even for human annotators. Therefore, we carefully curated prompts to ensure a one-to-one correspondence between prompts and physical laws, with clear physical phenomena demonstrating each law. By considering various physical laws across multiple scenarios, \bench can comprehensively assesses current T2V models' understanding of physical commonsense.

Multiple physical laws could be included in a single prompt, which may bring confusion to the evaluation of physical common sense in video generation, even for human annotators. To avoid this, we carefully curate prompts to ensure a one-to-one correspondence for each physical phenomenon it reflects, with clear physical law inside. By incorporating physical laws from four distinct physical categories, \bench offers a thorough assessment of current T2V models' understanding of physical commonsense.
% \begin{figure*}[htbp]
%   \centering
%   \scalebox{0.9}{
%   \includegraphics[width=\linewidth]{crop_construction.pdf}
%   }
%   % \vspace{-70pt}
%   \caption{
%   The \bench data pipeline covers $4$ physics categories. 
%   For each, we selected important physical laws and manually crafted initial prompts that reflect the corresponding physical phenomena. Then, GPT-4o is used to augment the initial prompts with more details and enhance the prompt diversity with different objects. In the end, we obtain $160$ T2V prompts after manual checking. We first augment the prompts to make them more detailed, then enhance the diversity by using GPT-4o to replace key objects. In the end, we obtain $160$ T2V prompts after manual checking. 
% %   \dianqi{Figure 3 should be merged with Figure 2 as Figure 2(c)}
%   }
%   \label{fig:construction}
%   % \vspace{-10pt}
% \end{figure*}
\paragraph{Benchmark Construction.} As shown in Figure \ref{fig:phybench} (b), we develop a comprehensive approach to create \bench. The methodology encompasses five steps:
\textbf{1) Conceptualization:} Following \citep{halliday2013fundamentals}, We first identify key physical commonsense from four major categories of physics. For each category, we select specific physical laws from textbooks \citep{harjono2020interactive}, which can be widely recognized and can be easily demonstrated through clear, observable physical phenomenon. 
% \dianqi{select? how do you select? from the reference?} that best represent these laws, prioritizing those that are widely recognized and can be easily demonstrated through clear, observable scenarios. The process involves brainstorming sessions informed by physics textbooks, online resources, and large language models.
\textbf{2) Prompt Engineering:} For each physical law, we manually craft the initial T2V prompts to clearly depict the underlying physical phenomenon
% while maintaining simplicity in visual composition. Each prompt is annotated with the corresponding physical law it represents.
\textbf{3) Prompt Augmentation:} To enhance the model's video generation capabilities, we augment the initial T2V prompts by adding additional details, such as more precise descriptions of objects and actions \citep{yang2024cogvideox}. This augmentation process is carefully designed to avoid revealing the expected physical phenomenon.
\textbf{4) Diversity Enhancement:} Following T2V-CompBench \citep{sun2024t2v}, we employ GPT-4o to perform object substitution on the augmented prompts. This step increases the diversity of the benchmark.
\textbf{5) Quality Control:} We conduct a thorough review of the prompts and their associated physical laws to ensure accuracy and relevance. Specifically, we ensure that the T2V prompts and corresponding physical laws are clear and accurate. We then randomly use the current T2V model
% from Section \ref{sec:evaluatedmodels} 
to check if the prompts are simple enough for the model to generate semantically accurate videos.
% \dianqi{what's the criteria for filtering in manual check?}.
This methodology yields a robust and comprehensive benchmark for assessing T2V models' comprehension of physical commonsense, providing a valuable tool for advancing research in this domain. For more detailed information about the dataset, please refer to the Appendix \ref{appendix:physgenbench}

\section{PhyGenEval}
\label{sec:eval}

\eval aims to assess whether the physical phenomena in the generated videos conform to the corresponding physical laws. To obtain a clear judgment, we decompose the evaluation into semantic alignment (SA) and physical commonsense alignment (PCA). While SA evaluates whether the semantic meaning inferred by the generated video and the input prompt are matched, PCA measures whether the evaluated physical laws are grounded in the videos. For example, for the scene \textit{``an egg collides with a stone"}, SA requires a video containing the egg, the stone, and the collision action. PCA necessitates a video for the whole physical motions in which the egg hits a stone and then breaks, while the stone remains intact. Following \citep{he2024mantisscore}, we convert both SA and PCA to a four-point scale, as well as the human ratings.

\begin{figure*}[t!]
  \centering
  \scalebox{1.0}{
  \includegraphics[width=\linewidth]{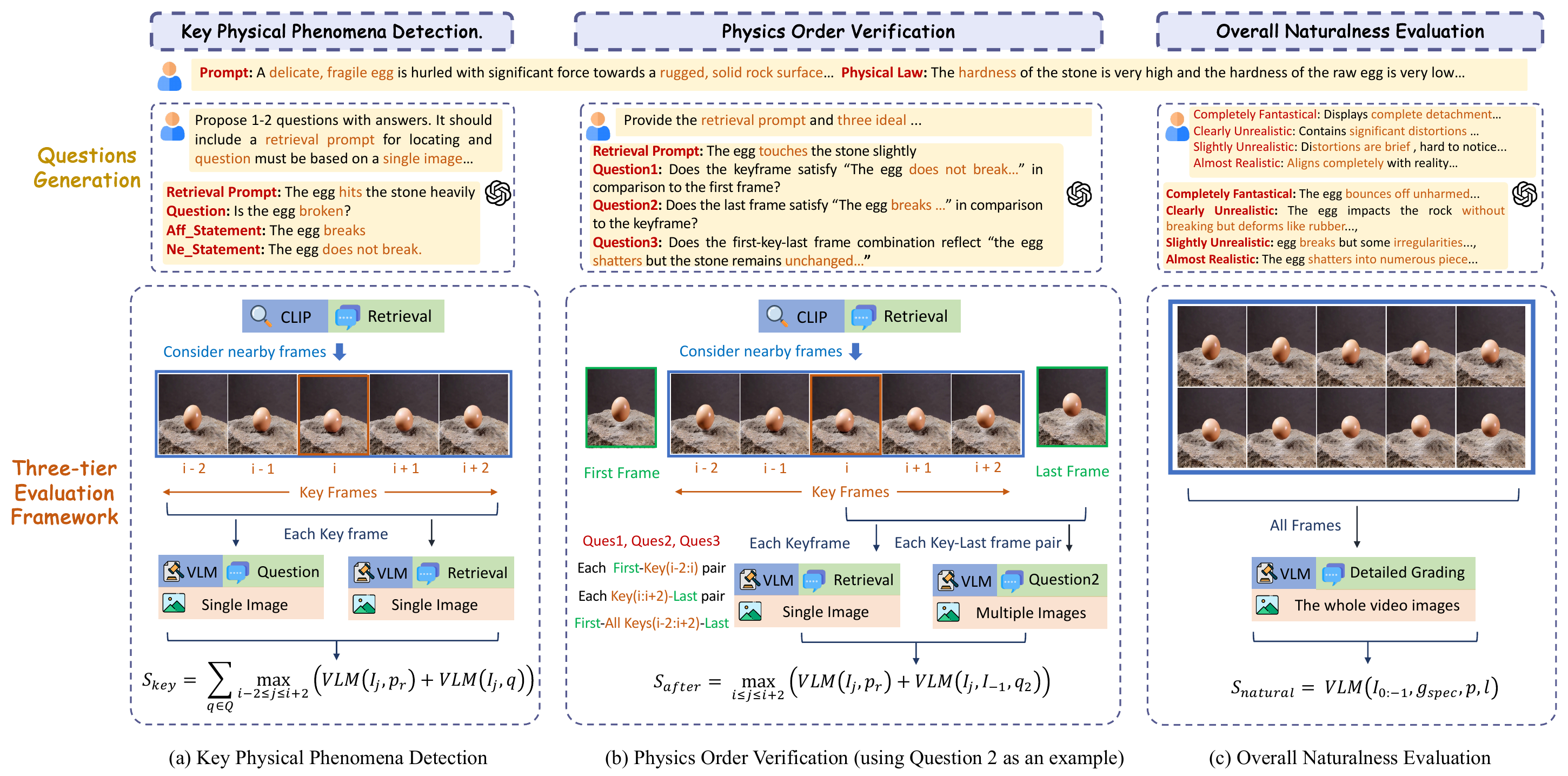}
  }
  % \vspace{-40pt}
  \caption{An overview of the proposed \eval. \eval is divided into three parts: Key Physical Phenomena Detection, Physics Order Verification, and Overall Naturalness Evaluation. Each part uses an appropriate VLM in combination with physical-based customized questions generated by GPT-4o. The final score is the combined result of the three parts. For the example in the figure, the three-stage scores are $0$, $1$ (only $q_1$ is correct), and $0$. The final score is calculated as 0 according to \ref{phyevaloverall}.}
  \label{fig:eval}
  \vspace{-10pt}
\end{figure*}

% \subsection{Semantic Alignment Evaluation}
% Evaluating text-video alignment  is a widely studied problem, with various methods using VLM as a human-aligned evaluator such as VideoScore \citep{he2024mantisscore}, and T2V-CompBench \citep{sun2024t2v}. Among them , VBench has significant limitations , as it only supports single objects and a limited range of actions. Our scenes, however, contain a wide variety of objects and actions. Therefore, we follow VideoScore and T2V-CompBench, using VLM for evaluation. However, these methods have several limitations: 1. The VLM models used in these approaches have limited capacity, leading to hallucinations. 2. When evaluating text-video alignment, prior methods often directly judge whether the object or action mentioned in the text prompt is present in the video. However, this approach can entangle action recognition with the correctness of the underlying physics, causing inaccuracies. To address these limitations in our benchmark, we made the following improvements: 1. We utilize GPT-4o~\citep{xx} as a more capable model. 2. We employ GPT to extract objects and actions from the original text prompt and then evaluate in stages. We first determine whether the object exists in the video, followed by assessing whether the action occurs. Experimental results have shown that our automated evaluation method achieves better results with high alignment to human judgment than previous methods. For more details, please see the appendix.

\subsection{Semantic Alignment Evaluation}
% Evaluating text-video alignment has been extensively investigated. 
% Previous methods \citep{he2024mantisscore, sun2024t2v} directly feed the prompt to the VLM and determine whether the described objects and actions appear. However, We find that these approaches do not generalize well to \bench because directly feeding the prompt intertwines semantic and physical correctness during evaluation, affecting the accuracy of the text-alignment assessment. 

Directly asking the Vision-Language Model(VLM)
% \dianqi{check, whether we have a definition for VLM previously} 
to align the semantic meaning between videos and input prompts are difficult, as prompts usually are mixed with semantic entities and physical phenomena, and the intermediate outcomes are subtly implied by the videos. For example, in a prompt like \textit{``A timelapse captures the transformation of soup as the temperature rises above 100°C"}, a possible video generation would appear like \textit{``The video shows a soup, but there is no transformation of the soup"}. To address the challenge, we first employ GPT-4o to extract object and action from the original text prompt, we then utilize GPT-4o to sequentially determine the presence of extracted objects in the video and verify the occurrence of specified actions.
% While the transformation should be evaluated under physical commonsense alignment, semantic alignment requires to evaluate 
% leading to confusion in the semantic alignment evaluation process. 
% We first address the challenges in semantic alighment, 
% We adopt a two-stage strategy. Initially, W. Subsequently,  
This decomposition provides more fine-grained captures and prevents the model from confusing semantic and physical correctness during evaluation. Experimental results demonstrate that our automated evaluation method aligns more closely with human judgment and outperforms previous methods \citep{he2024mantisscore, sun2024t2v} in \bench (Appendix~\ref{appendix:SAdetail}).

% \dianqi{by reading this paragraph, I still don't understand how we evaluate the SA, even though I know how you evaluate them. }

% For more details, please see the appendix.

% However, these methods have several limitations: 1. The VLM models used in these approaches have limited capacity, leading to hallucinations. 2. When evaluating text-video alignment, prior methods often directly judge whether the object or action mentioned in the text prompt is present in the video. However, this approach can entangle action recognition with the correctness of the underlying physics, causing inaccuracies. To address these limitations in our benchmark, we made the following improvements: 1. We utilize GPT-4o~\citep{xx} as a more capable model. 2. We employ GPT to extract objects and actions from the original text prompt and then evaluate in stages. We first determine whether the object exists in the video, followed by assessing whether the action occurs. Experimental results have shown that our automated evaluation method achieves better results with high alignment to human judgment than previous methods. For more details, please see the appendix.

\subsection{Physical Commonsense Evaluation}\label{phyevalpc}
% Although some studies suggest that VLMs serve as human-aligned evaluators for text-to-image or text-to-3d generation models \citep{lin2024evaluating,wu2024gpt}, we find that directly using VLMs struggles to analyze whether videos conform to physical laws. We attribute this to: i) videos have an additional temporal dimension compared to images, making the task more challenging; ii) VLMs for videos are generally weaker than those for images, and the performance of the models deteriorates as the number of input frames increases to a certain extent \citep{li2024llava,ye2024mplug}; iii) directly using VLMs makes it difficult to comprehend the embedded physical knowledge.
% Recent studies indicate that VLMs can serve as human-aligned evaluators for text-based generative tasks \citep{lin2024evaluating, wu2024gpt}. 

% Recent studies have shown that VLMs can also be employed as evaluators in video generation to assess the quality of dynamic processes \citep{lin2024evaluating} and the correctness of spatial relationships \citep{sun2024t2v} in videos. However, \dianqi{too verbose, just say we evaluated multiple common used benchmarks and discovered them have poor performance correlated with human assessments}these evaluators achieve poor performance in assessing video conformity with  as shown in Table \ref{tab:correlation}\dianqi{table 1 can put to appendix}, 

To evaluate physical correctness in the video, we evaluated multiple common evaluation metrics comparing human assessments\footnote{Annotators are asked to score the correctness of physical commonsense in the video. Details refer to Section \ref{sec:evaluatedmodels} and Appendix \ref{appendix:humaneval}}. Experimental results in Table~\ref{tab:correlation}
% \dianqi{table 1 can put to appendix} 
demonstrate that these methods struggle to generalize to the assessment of physical commonsense correctness on \bench, e.g., VideoScore \citep{he2024mantisscore} has only a spearman correlation of $0.19$ on \bench, which is most correlated with human assessments except \eval. We attribute it to the main factor: Directly using video-based VLMs fails to comprehend the embedded physical commonsense \citep{jassim2023grasp}, as current methods are not designed with physical commonsense as a foundation. To fully understand the physical commonsense in the video, there are three key factors need to solve: \textbf{i): }Physical processes typically exhibit clear key phenomena depicted by the input prompt (e.g., \textit{``the egg breaks upon hitting the rock."}). It is necessary to identify these key physical phenomena and detect their presence in videos.
\textbf{ii): }Physical processes are characterized by causality, manifested in the correct sequence of critical events(e.g., \textit{``The egg touchs the rock first, then breaks."}). The correct sequence order validates the correctness of physical processes.
\textbf{iii): }Physical processes need to possess overall naturalness, which represents the realistic of the overall process. 
% Some studies directly using the video-based VLM like DEVIL \citep{liao2024evaluation} struggle to handle this aspect effectively (Table \ref{tab:correlation}).
To address these factors, we design a progressive strategy that starts with key physical phenomena, then moves through the sequence of several key phenomena, and finally evaluates the overall naturalness of the entire video. This hierarchical and refined approach reduces the difficulty compared to existing methods that directly uses VLMs to evaluate physical commonsense, enabling \eval to achieve results closely aligned with human judgements.

\paragraph{Key Physical Phenomena Detection. } This stage aims to detect \textsl{whether the key physical phenomena occur in the video}. Here we define the key phenomena as an observable and distinctive occurrence (e.g., a specific frame) within a physical process that can directly reveal the corresponding physical law, like deformations or color changes. 
% For instance, it could be that an egg cracks upon hitting a stone. 
% which could be that an egg cracks upon hitting a stone. 
% such as identifying the where key phenomena that an egg cracks upon hitting a stone.
% \dianqi{for example, there is no motivation and why we need to do this step}. % To this end, we first obtain a retrieval prompt $P_r$ and physical-related questions $Q$ \dianqi{from high-level, what's the purpose of retrieval prompt and physical-related questions?} specific to each prompt in our \bench by prompting GPT-4o with input prompt and underlying physical laws as shown in Figure \ref{fig:eval} (a). The retrieval prompt is used to locate the key frame where the physical phenomena happen. And physical-related questions are utilized to check whether the expected physical phenomena are present in the key frame. 
% To achieve this,
% \dianqi{xinyu:motivation of the design should be ahead, retrieval prompt $P_r$ and question $Q$ can be separated,first introduce $P_r$,for example,To locate the key phenomena frame,we designed a retrieval prompt $P_r$ ...To check whether the expected physical phenomena are present,we designed question  $Q$+ figure link}
For each input prompt in \bench, we craft a retrieval prompt $p_r$ and a set of physics-related questions $Q$, where the retrieval prompt is used to locate the key phenomena frame, and physical-related questions are utilized to check whether the expected physics phenomena are present in the keyframe. 
% \dianqi{explain why we need multiple questions for each prompt in here}. 

% \dianqi{check how I changed the order and make the logic clear.}

As illustrated in Figure \ref{fig:eval} (a), we first obtained both $Q$ and $P_r$ by prompting GPT-4o with the input T2V prompt and corresponding physical law. Following \citep{hessel2021clipscore}, a keyframe $I_{i}$ from the video based on the retrieval prompt, where $I_i$ is the $i$-th frame in the video. By using the keyframe, we define a confidence score of the key phenomena in the video:. 
% Formally, for given retrieval prompt $P_r$ and question $Q$, it is given by
% we define the score $\mathrm{Score_{0}}$ in single image evaluation for the first given question by 
\begin{equation*}
\label{eq:singlescore}
    % L(\mathbf{x}, \mathbf{b}, y) = \prod_{k=1}^{m}  P_{\mathrm{CMLM}}\left({z}_{k} \mid \boldsymbol{\tilde{\mathbf{x}}}_{\mathbf{b} \backslash z_k} ; \Theta_\textit{y}\right).
    \mathrm{S_{key}} = \sum_{q \in Q}\max_{i-2 \leq j \leq i+2} \left(\mathrm{VLM}(I_j, q) + \mathrm{VLM}(I_j, p_r)\right),
\end{equation*}
% $\mathrm{Score_{0}} = \max_{i-2 \leq j \leq i+2} \left( \mathrm{VLM}(\mathrm{Img}_j, Q) + \mathrm{VLM}(\mathrm{Img}_j, P_r) \right)$ 
where $\mathrm{VLM}(I_j, q)$ reflects the presence of physical phenomena in $I_j$ for each related question $q$ from $Q$. $\mathrm{VLM}(I_j, p_r)$ checks whether $I_j$ matches the retrieval prompt,
% \dianqi{why we need this term? shouldn't $I$ retrieved by $p$}, 
which ensures key phenomena occur at the correct frame. Since videos may contain semantic errors, it's also important for determining if key physical phenomena occur (e.g., an egg shouldn't break in mid-air before hitting a rock).
We consider adjacent $5$ frames near the keyframe to enhance the robustness. For example, the egg may not be cracked just when it first contacts the stone. We instantiate VLM-based evaluator $\mathrm{VLM}(\cdot)$ with VQAScore \citep{lin2024evaluating}, which has been shown promising evaluation results on visual question-answering.

\paragraph{Physics Order Verification. } In this stage, we verify \textsl{whether key physical phenomena occur in the correct order}. The correct physical sequence is an ordered series of events in a physical process that reflects causality, which represents the necessary prerequisites and temporal order of key physical phenomena. As an example, the egg should first touch the stone and then crack. Considering current models in \bench generally maintain outcome consistency \citep{huang2024vbench} (e.g., the egg would not reassemble itself after it is broken). we approach this direction by investigating the order correctness from the keyframes (Figure \ref{fig:eval} (b)), e.g., the keyframe of the egg hits the stone should be ahead of the keyframe of the broken egg.

% from the first frame to the key frame, and from the key frame to the last frame, leading to multiple multi-image QA tasks.
% In this stage, we focus on verifying \textbf{whether the sequence of key physical phenomena occurs correctly}. For instance, when an egg cracks upon hitting a stone, the egg should first impact the stone, and then break. We observe that existing models generally maintain subject consistency in \bench (e.g., the egg does not break and then reassemble itself.). This observation also aligns with the conclusions drawn from VBench \citep{huang2024vbench}. Therefore, we aim to approach this goal using a multi-image question-answering task, which is based on the first and last key frame pair.
% Considering that the current T2V model possesses temporal and subject consistency \citep{huang2024vbench}, we approach this goal using a multi-image question-answering task, which is based on the first and last key frames.

Similar to the single image evaluation, we prompt GPT-4o to generate a retrieval prompt $p_r$ and three physical-related questions $(q_1, q_2, q_3)$. $p_r$ is used to locate the keyframe (e.g., the moment the egg slightly touches the stone.). While $q_1$, $q_2$, and $q_3$ are questions to check the order correctness from the first frame to the keyframe, from the keyframe to the last frame, and from the first frame to the last frame, respectively. 
%
%targeting the first-key, key-last, and first-key-last frame pairs respectively. These questions are used to verify the correctness of critical sequences (e.g., the egg should not break in the first-key frame pair but break in the key-last frame pair).
%
Similarly, we first use CLIPScore to locate the key frame $I_{i}$, then the order correctness scores of $\mathrm{S_{before}}$ and $\mathrm{S_{after}}$ are defined as:
%
%similar to the single image stage, which is based on the first-key frame pair and key-last frame pair respectively.
% \begin{equation*}
% \label{eq:likelihood}
%     L(\mathbf{x}, \mathbf{b}, y) = \prod_{k=1}^{m}  P_{\mathrm{CMLM}}\left({z}_{k} \mid \boldsymbol{\tilde{\mathbf{x}}}_{\mathbf{b} \backslash z_k} ; \Theta_\textit{y}\right).
% \end{equation*}
% \dianqi{check my equation style, and use the similar styles}
% \begin{equation*}
\begin{align*}
\label{eq:multiscore}
    \mathrm{S_{before}} = \max_{i-2 \leq j \leq i} \left( \mathrm{VLM}(I_0, I_j, q_1) + \mathrm{VLM}(I_j, p_r) \right) \\
    \mathrm{S_{after}} = \max_{i \leq j \leq i+2} \left( \mathrm{VLM}(I_j, I_{-1}, q_2) + \mathrm{VLM}(I_j, p_r) \right)
\end{align*}
% \begin{equation}
%     \mathrm{Score_{1}} = \max_{i-2 \leq j \leq i} \left( \mathrm{VLM}(\mathrm{Img}_0, \mathrm{Img}_j, Q_1) + \mathrm{VLM}(\mathrm{Img}_j, P_r) \right)
% \end{equation} 
% %
% \begin{equation}
%     \mathrm{Score_{2}} = \max_{i \leq j \leq i+2} \left( \mathrm{VLM}(\mathrm{Img}_j, \mathrm{Img}_{-1}, Q_2) + \mathrm{VLM}(\mathrm{Img}_j, P_r) \right)
% \end{equation}
%
$q_3$ assesses the overall physical sequence coherence of the video. The score of answering $q_3$ is defined as by $\mathrm{S_{all}} = \mathrm{VLM}(I_0, I_{i-2: i+2}, I_{-1}, q_3)$, 
which evaluates the overall sequence (similar to the input video but using manually selected key frames).
% \dianqi{why $Score_3$ is necessary?need further explanation}. 
Here we employ GPT-4o or LLaVA-Interleave \citep{li2024llava} as the VLM-based evaluator $\mathrm{VLM}(\cdot)$, as they demonstrate exceptional multi-image comprehension capabilities. The overall score of whole physical order evaluation can be formulated as $\mathrm{S_{order}} = \mathrm{S_{before}} + \mathrm{S_{after}} + \mathrm{S_{all}}$

\paragraph{Overall Naturalness Evaluation. }\label{videoeval} This stage aims to evaluate\textbf{ the overall naturalness of the video}. we define naturalness as the dynamic progression that aligns with real-world physical phenomenons \citep{liao2024evaluation}. For each prompt in \bench, we obtain a naturalness evaluation standard, denoted as $g_{spec}$, which is used to assess the naturalness for video. As shown in Figure \ref{fig:eval} (c), we first refer to DEVIL \citep{liao2024evaluation} to establish a general evaluation standard: $g_{gen}$, applicable to all T2V prompts. Besides, we use each input T2V prompt $p$, the corresponding physical law $l$, and general evaluation standard $g_{gen}$ to guide GPT-4o in generating a detailed evaluation standard: $g_{spec}$, for the given prompt. Finally, we require the VLM to score based on $p$, $l$, $g_{spec}$, and the corresponding video denoted by $I_{0:-1}$. Formally, we define the overall naturalness score as:
\begin{equation*}
\label{eq:naturalscore}
    % L(\mathbf{x}, \mathbf{b}, y) = \prod_{k=1}^{m}  P_{\mathrm{CMLM}}\left({z}_{k} \mid \boldsymbol{\tilde{\mathbf{x}}}_{\mathbf{b} \backslash z_k} ; \Theta_\textit{y}\right).
    \mathrm{S_{natural}} = \mathrm{VLM}(I_{0:-1}, p, l, g_{spec})
\end{equation*}
We implement the VLM-based evaluator $\mathrm{VLM}(\cdot)$ using InternVideo2 \citep{wang2024internvideo2} and GPT-4o, both of which have demonstrated promising results in video understanding.

\paragraph{Overall Score. }\label{phyevaloverall}  We first discretize $\mathrm{S_{key}}$, $\mathrm{S_{order}}$, and $\mathrm{S_{natural}}$ from the three stages into a four-point scale, then take their average and apply floor rounding as the final score. For robust purposes, we evaluate $\mathrm{S_{order}}$ with both GPT4o and LLaVA-Interleave and $\mathrm{S_{natural}}$ with both GPT4o and InternVideo2. The final score is calculated as the ensemble of two methods. Detailed calculation protocols are provided in Appendix~\ref{appendix:PhysEval}.

\section{Experiment}

% \subsection{Experiments Setup}

% \textbf{Evaluation metric. } We provide the average scores for scene and physical correctness assessments by \eval. To better measure the alignment between machine scoring and human scoring, we calculate the Kendall coefficient and Spearman correlation coefficient based on the machine and human evaluations for each sample. 

\paragraph{Experiments Setup.}\label{sec:evaluatedmodels} We evaluate $5$ open-source models including OpenSora V1.2 \citep{opensora}, Lavie \citep{wang2023lavie}, CogVideoX 2b \citep{yang2024cogvideox}, CogVideoX 5b \citep{yang2024cogvideox}, and Vchitect2.0 \citep{wang2023lavie}, as well as proprietary models Kling \citep{kling}, Pika \citep{Pika}, and Gen-3 \citep{gen3}. We compare our proposed metric with existing metrics or benchmarks: Videophy \citep{bansal2024videophy}, VideoScore \citep{he2024mantisscore} and DEVIL \citep{liao2024evaluation}
% 1) Videophy \citep{bansal2024videophy} uses a video understanding model to assess whether the generated video adheres to dynamic motion correctness. 2) VideoScore \citep{he2024mantisscore} functions like a video version of VQAScore \citep{lin2024evaluating}, providing the score for the naturalness of the video content. 3) DEVIL \citep{liao2024evaluation} directly employs Gemini Pro 1.5 \citep{reid2024gemini} to rate the naturalness of videos based on a consistent template. 
More Detailed information is provided in Appendix~\ref{appendix:experiment}. 

For human evaluation, we compared the results across $8$ T2V models. We randomly select $64$ prompts from \bench and generate $64$ videos for each T2V model. Therefore we need evaluation $512$ videos. We ask three annotators to provide semantic and physical scores for each video\footnote{Note that we ask the annotators to focus on the correctness of the physical phenomena for physical scores.}. Each annotator will give an integer score of 0-3 for the semantic and physical scores, and the final score is the average of the three scores and rounded up. Finally, we calculate the correlation between the human scores and automatic evaluation scores using Kendall’s $\tau$ and Spearman’s $\rho$. we pue more detailed information about human evaluation in Appendix \ref{appendix:humaneval}.

% \subsection{Human Evaluation Comparing Existing Benchmarks} 

% Here we use GPT-4o. 4) ChronoMagicBench \citep{} utilizes a video retrieval model such as InternVideo2 \citep{} with a unified template to score the dynamic properties of videos; here, we replace the template with the one used in DEVIL to also score the naturalness of the video.

% \paragraph{Human evaluation. }

\paragraph{Human Evaluation. }\label{main:humaneval} As shown in Table \ref{tab:correlation}, current video generation evaluation metrics largely overlook physical correctness. In contrast, \eval implements a detailed design for evaluating physical correctness, demonstrating strong correlations with human judgments across all categories. Its overall correlation coefficient reaches $0.81$, indicating that \eval serves as an effective human-aligned physical commonsense correctness evaluator for \bench. We put more results in Appendix \ref{appendix:quanti}

\begin{figure*}[t!]
  \centering
  \scalebox{0.9}{
  \includegraphics[width=\linewidth]{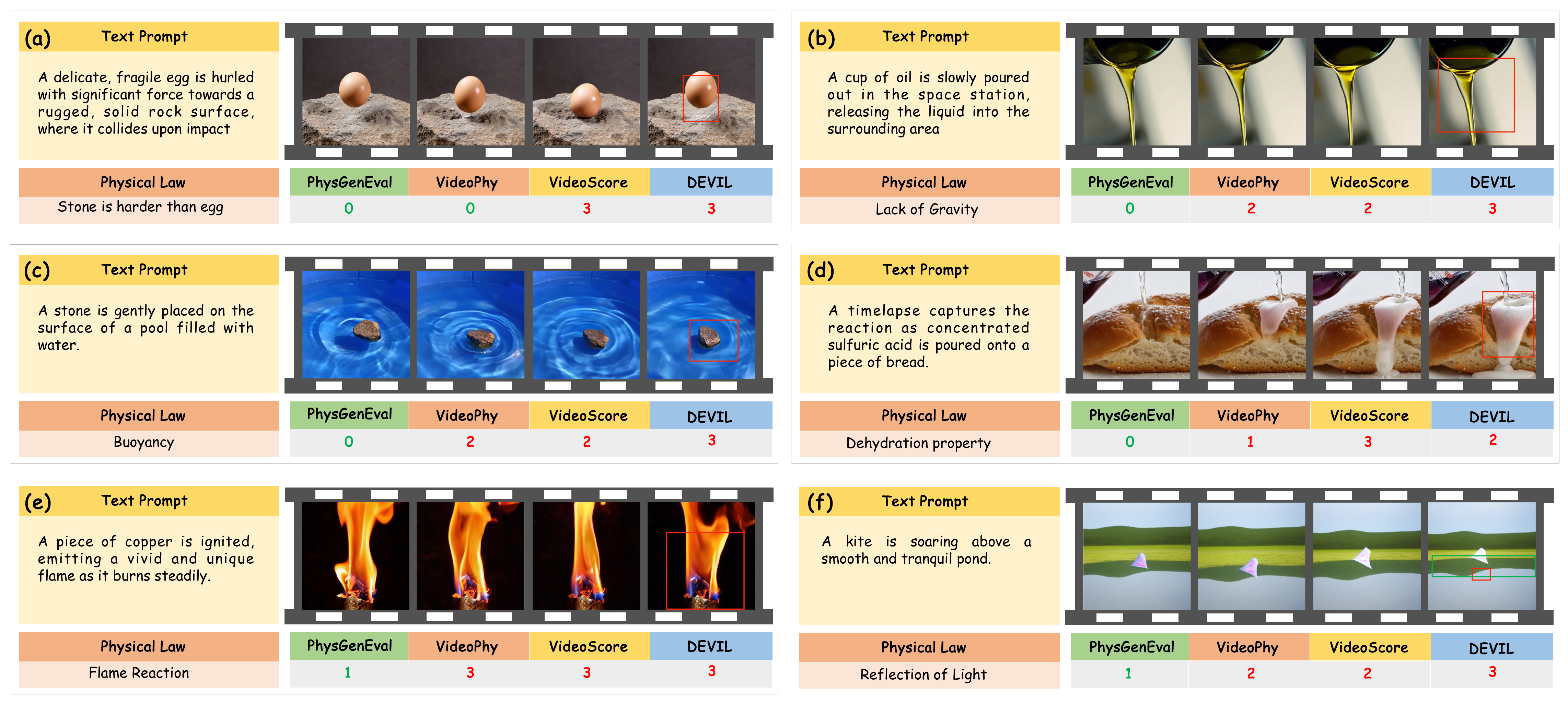}
  }
  \caption{Different video generation evaluation metric in \bench. Except for the proposed \eval, the current methods cannot reasonably assess the correctness of physical commonsense in videos from \bench.}
  \label{fig:metric}
  % \vspace{-10pt}
\end{figure*}

We conduct several case studies to illustrate the differences between various metrics more clearly. As shown in Figure \ref{fig:metric}, (a) and (f) reveal that VideoScore and DEVIL are prone to misclassifying videos that have smooth and consistent motion but violate fundamental physical laws. Specifically, as for (a), when \textit{``an egg exhibits rubber-like elasticity upon impact with a rock instead of breaking,"} these metrics incorrectly evaluate it as physically correct. VideoPhy exhibits similar limitations. In (c), it incorrectly assesses \textit{``a rock floating on water instead of sinking"} as physically correct. Furthermore, our analysis reveals a major flaw in these three methodologies: they cannot incorporate domain-specific physical commonsense. As illustrated in (e), where \textit{``the flame from burning copper appears red instead of green,"} these metrics fail to identify the mistake. This indicates their inability to incorporate domain-specific physical commonsense. In contrast, \eval demonstrates a robust integration of physical commonsense and comprehensive video content analysis, resulting in more accurate and physically consistent evaluations in \bench.
% While VideoScore and DEVIL include measures of video naturalness, they tend to focus more on video fluidity and neglect the accuracy of physical phenomena, resulting in poor correlation with human judgments. Although VideoPhy also considers physical correctness, its correlation with human ratings remains weak. We attribute this to VideoPhy's input not including the corresponding prompt for the video, thus failing to account for specific scenarios and environmental conditions.  To facilitate comparison between different methods, we provide a visualization of evaluation results from various approaches in Figure x. We put more results in the section xx in the appendix.
% \begin{table}
% \resizebox{\linewidth}{!}{%  
% \begin{tabular}{l cccc cccc cc}
% \toprule
% \multirow{2}{*}{ \textbf{Metric}} & \multicolumn{2}{c}{\textbf{Mechanics}} & \multicolumn{2}{c}{\textbf{Optics}} & \multicolumn{2}{c}{\textbf{Thermal}} & \multicolumn{2}{c}{\textbf{Material}} & \multicolumn{2}{c}{\textbf{Overall}} \\
% \cmidrule(lr){2-3}\cmidrule(lr){4-5}\cmidrule(lr){6-7}\cmidrule(lr){8-9}\cmidrule(lr){10-11}\cmidrule(lr){12-13} \\

% &  & $\rho$($\uparrow$)& $\tau$($\uparrow$)  & $\rho$($\uparrow$)& $\tau$($\uparrow$) & $\rho$($\uparrow$) & $\tau$($\uparrow$) & $\rho$($\uparrow$) & $\tau$($\uparrow$) \\
%     \midrule
%     DEVIL   \\
%     ChronoMagicBench  \\
%     VideoPhy   \\
%     VideoScore   \\
%     \eval   \\
%     \bottomrule
% \end{tabular}
% }
% \label{tab:human_corr}
% \end{table}
\begin{table}
\centering
    \caption{\textbf{PCA correlation results with proposed \eval in video generation}.  \eval is significantly closer to human feedback on \bench compared to other metrics. 
} 
\label{tab:correlation}
\resizebox{\linewidth}{!}{%  
\begin{tabular}{l cccc cccc cc}
\toprule
\multirow{2}{*}{ \textbf{Metric}} & \multicolumn{2}{c}{\textbf{Mechanics}} & \multicolumn{2}{c}{\textbf{Optics}} & \multicolumn{2}{c}{\textbf{Thermal}} & \multicolumn{2}{c}{\textbf{Material}} & \multicolumn{2}{c}{\textbf{Overall}}\\
\cmidrule(lr){2-3}\cmidrule(lr){4-5}\cmidrule(lr){6-7}\cmidrule(lr){8-9}\cmidrule(lr){10-11}
 & $\tau$($\uparrow$) & $\rho$($\uparrow$) & $\tau$($\uparrow$) & $\rho$($\uparrow$) & $\tau$($\uparrow$) & $\rho$($\uparrow$)& $\tau$($\uparrow$) & $\rho$($\uparrow$) & $\tau$($\uparrow$) & $\rho$($\uparrow$)  \\
    \midrule
    DEVIL  \citep{liao2024evaluation}  & $0.15$ & $0.16$ & $0.03$ & $0.03$ & $0.10$ & $0.11$ & $0.27$ & $0.29$ & $0.17$ & $0.18$\\
    VideoPhy \citep{bansal2024videophy}  & $0.00$ & $-0.03$ & $-0.15$ & $-0.14$ & $0.08$ & $0.08$ & $0.13$ & $0.14$ & $0.03$ & $0.04$\\
    VideoScore \citep{he2024mantisscore}  & $0.18$ & $0.20$ & $0.07$ & $0.08$ & $0.14$ & $0.15$ & $0.14$ & $0.15$ & $0.17$ & $0.19$\\
    \eval  & $\textbf{0.72}$ & $\textbf{0.75}$ & $\textbf{0.76}$ & $\textbf{0.77}$ & $\textbf{0.73}$ & $\textbf{0.75}$ & $\textbf{0.81}$ & $\textbf{0.84}$ & $\textbf{0.78}$ & $\textbf{0.81}$\\
    \bottomrule
\end{tabular}
}

\end{table}
\paragraph{Quantitative Evaluation.} We conduct extensive experiments on a wide range of popular video generation models. 
% As shown in Table \ref{tab:sa-benchmark} in the appendix, nearly all models achieve relatively high SA scores, regardless of whether the evaluation is conducted by machines or humans. This indicates that the scenarios in \bench are relatively simple, facilitating the evaluation of physical commonsense\dianqi{didn't get the logic, are you want to say the semantic meaning in the constructed physical scenario for evaluation is quite simple? }. 
As illustrated in Table \ref{tab:pca-benchmark}, even the best-performing model, Gen-3, only attains a PCA score of $0.51$ on \bench. This indicates that even for prompts containing obvious physical commonsense, current T2V models struggle to generate videos that comply with intuitive physics. It indirectly reflects that these models are still far from achieving the world simulator.

% Furthermore, we identified the following key observations for physical commonsense understanding: \textbf{i):} Across different categories of physical commonsense, all models consistently score higher in the domain of optics compared to mechanics, thermodynamics, and other areas. Notably, Vchitect2.0 achieves a physics score in the optics domain comparable to that of closed-source models. We hypothesize that the superior performance in the optics domain can be attributed to the abundant and explicit representation of optical knowledge in pre-training datasets, which enhances the model's comprehension in this area. \textbf{ii):} Kling and Gen-3 exhibit significantly better performance than other models. Specifically, Gen-3 demonstrates a strong understanding of material properties, achieving a score of $0.51$, which is substantially higher than other models. Kling performs relatively well in thermodynamics, with the highest score of $0.50$ in this domain. \textbf{iii):} Among open-source models, Vchitect2.0 and CogVideoX 5b perform relatively well, both approaching the performance level of Pika. In contrast, Lavie consistently exhibits lower physical correctness across all categories.

Furthermore, we identify the following key observations: \textbf{1): }Across various categories of physical commonsense, all models consistently demonstrate superior performance in the domain of optics compared to other areas. Notably, Vchitect2.0 and CogVideoX-5b achieve a PCA score in the optics domain comparable to that of closed-source models. We posit that this superior performance in the optics domain can be attributed to the abundant and explicit representation of optical knowledge in pre-training datasets, thereby enhancing the model's comprehension in this area. \textbf{2): }Kling and Gen-3 exhibit significantly higher performance compared to other models. Specifically, Gen-3 demonstrates a robust understanding of material properties, achieving a score of $0.51$, which substantially surpasses other models. Kling performs particularly well in thermal, attaining the highest score of $0.50$ in this domain. \textbf{3): }Among open-source models, Vchitect2.0 and CogVideoX 5b perform comparatively well, both exceeding the performance level of Pika. In contrast, Lavie consistently exhibits lower physical correctness across all categories.

\begin{table*}[t]
\centering
    \caption{\textbf{Evaluation results of PCA with the proposed \eval in videos generated by several models }. The results reveal that all models score very low in physical commonsense accuracy, highlighting that current T2V models face significant challenges in correctly grasping physical commonsense.
} 
\label{tab:pca-benchmark}
\resizebox{\linewidth}{!}{%
\begin{tabular}%{lcccccccccc}
{lccccccccccccccccccc}
\toprule %\multirow{2}{*}
\multicolumn{1}{c}{{\textbf{Model}}} & \multicolumn{1}{c}{{\textbf{Size}}} & \multicolumn{1}{c}{{\textbf{Mechanics}}($\uparrow$)} & \multicolumn{1}{c}{\textbf{Optics}($\uparrow$)} & \multicolumn{1}{c}{\textbf{Thermal}($\uparrow$)} & \multicolumn{1}{c}{\textbf{Material}($\uparrow$)} & \multicolumn{1}{c}{\textbf{Average}($\uparrow$)} & \multicolumn{1}{c}{\textbf{Human}($\uparrow$)}    \\
\midrule

CogVideoX \citep{yang2024cogvideox} & 2B  & $0.38$  & $0.43$  & $0.34$  &  $0.39$  &  $0.39$ &  $0.31$ \\
CogVideoX \citep{yang2024cogvideox} & 5B  & $0.39$  & $0.55$  & $0.40$  &  $0.42$  & $0.45$  &  $0.37$ \\
Open-Sora V1.2 \citep{opensora} & 1.1B & $0.43$  & $0.50$  & $0.44$  &  $0.37$  &  $0.44$  &  $0.35$ \\
Lavie \citep{wang2023lavie} & 860M & $0.30$  & $0.44$  & $0.38$  &  $0.32$  &  $0.36$ &  $0.30$ \\ 
Vchitect 2.0 \citep{wang2023lavie} & 2B & $0.41$  & $0.56$  & $0.44$  &  $0.37$  &  $0.45$ &  $0.36$ \\ 
\midrule
Pika \citep{Pika} & - & $0.35$  & $0.56$  & $0.43$  &  $0.39$  &  $0.44$ &  $0.36$ \\
Gen-3 \citep{gen3} & - & $\textbf{0.45}$ & $0.57$  & $0.49$  &  $\textbf{0.51}$  &  $\textbf{0.51}$ &  $\textbf{0.48}$  \\
Kling \citep{kling} & - & $\textbf{0.45}$  & $\textbf{0.58}$  & $\textbf{0.50}$ &  $0.40$  &  $0.49$ &  $0.44$ \\

\bottomrule
\end{tabular}
}
\end{table*}

\begin{figure*}[t!]
  \centering
  \scalebox{0.9}{
  \includegraphics[width=\linewidth]{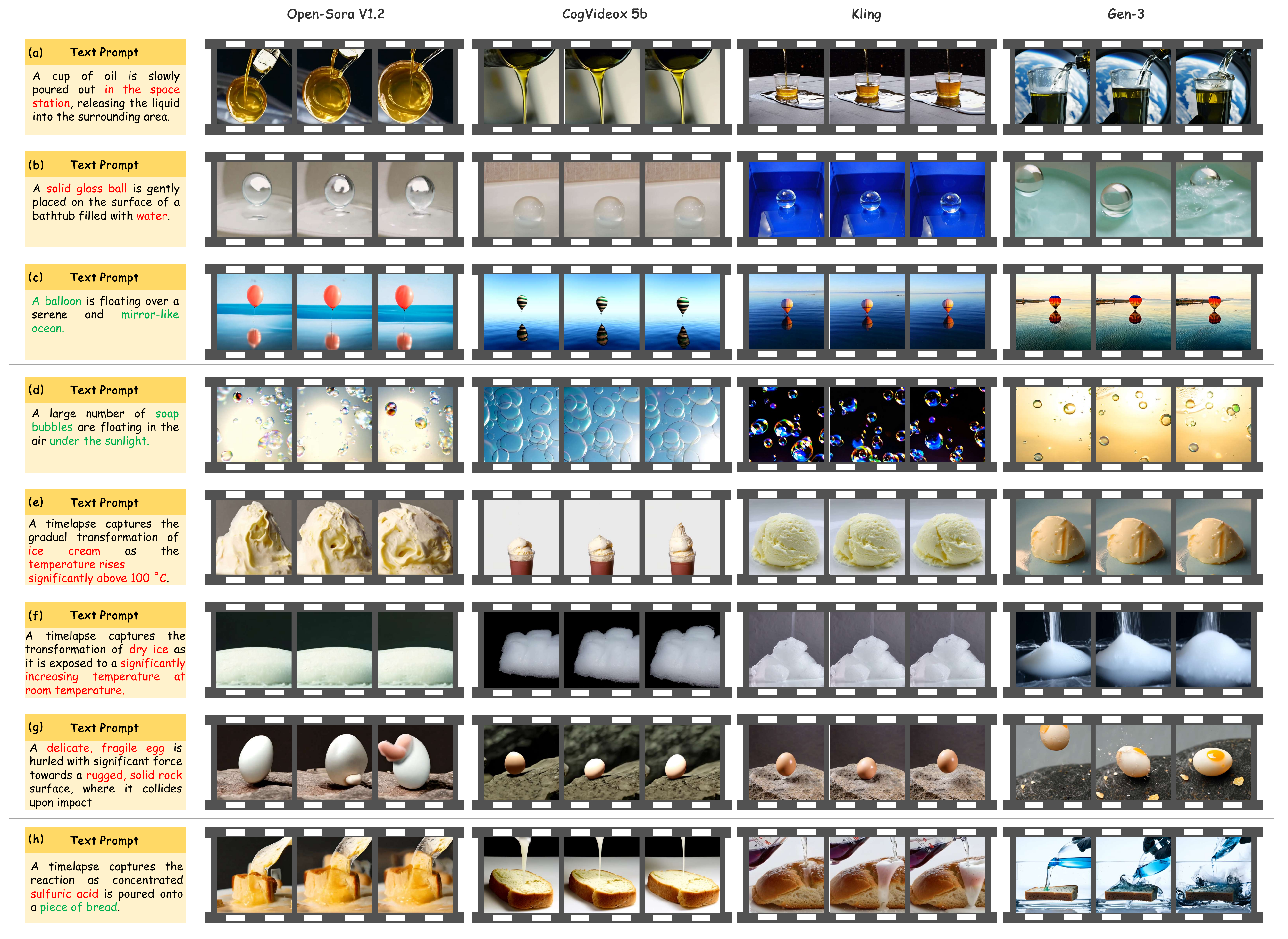}
  }
  % \vspace{-70pt}
  \caption{Qualitative comparisons of four categories. Current models perform relatively well in generating optical phenomenons but are weaker in mechanics, thermal, and material properties.}
  \label{fig:qualitive}
  \vspace{-15pt}
\end{figure*}

\paragraph{Qualitative Evaluation. } The different video cases for $4$ physical commonsense categories are illustrated in Figure \ref{fig:qualitive}. Our main observations are as follows:
In mechanics, the models struggle to generate simple physically accurate phenomenons. As shown in Figure \ref{fig:qualitive}, all models fail to depict the glass ball sinking in water. As for (b), instead showing it floating on the surface, OpenSora and Gen-3 even produce videos where the ball is suspended. Additionally, the models do not capture special physical phenomenonss, such as the state of water in zero gravity, as seen in (a). In optics, the models perform relatively better. (c) and (d) show the models handling reflections of balloons in water and colorful bubbles, though OpenSora and CogVideoX still produce reflections with noticeable distortions in (d). In thermal, the models fail to generate accurate videos of phase transitions. For the melting phenomenon in (e), most models show incorrect results, with CogVideoX even producing a video where the ice cream increases in size. Similar errors appear in the sublimation process in (f), with only Gen-3 showing partial understanding. Regarding material properties, (g) shows all models failing to recognize that an egg should break when hitting a rock, with Kling displaying the egg bouncing like a rubber ball. For simple chemical reactions, such as the black bread experiment in (h), none of the models demonstrate an accurate understanding of the expected reaction.

\vspace{-0.05in}

\paragraph{Ablation Study. } We conduct a detailed robustness analysis of the design elements in PhyGenEval, including the role of each level in the three-tier evaluation framework and the impact of the two-stage strategy proposed in overall naturalness evaluation. Experimental results show that the key designs of \eval are essential. Detailed results are provided in Appendix \ref{appendix:abli}.

\vspace{-0.05in}

\section{Discussion} To explore potential solutions for the challenges posed by \bench, We focus on widely used and proven-effective methods such as scaling laws \citep{kaplan2020scaling}, prompt engineering \citep{fu2024commonsense}, and some method like Venhancer \citep{he2024venhancer} aimed to improve general video quality \citep{huang2024vbench}. And we determine whether they can resolve the inability of current T2V models to generate videos aligned with physical commonsense. Through quantitative and qualitative analysis, we find: 1) Scaling up models can solve some issues but still fails to handle dynamic physical phenomenons, which we believe requires extensive training on synthetic data. 2) Prompt engineering like \citep{fu2024commonsense} only solves a few simple issues (e.g., flame color), highlighting the difficulty and significance of \bench. 3) While some methods improve general video quality, they do not enhance the model's understanding of physical commonsense. More detailed results are provided in Appendix \ref{appendix:discussion}.

\section{Conclusion}
In this paper, we explore the gap between current T2V models' understanding of physical commonsense and their role as world simulators. To achieve this, we introduce \bench and \eval. \bench is a benchmark specifically designed to assess models' understanding of physical commonsense, featuring various physical laws and simple, clear physical phenomenons. Alongside \bench, we propose a novel three-tier hierarchical evaluation framework called \eval to automate the evaluation process. Experimental and analytical results show that current T2V models struggle to generate videos that align with physical commonsense, highlighting a significant gap from world simulation. Moreover, simply scaling up models or applying prompt engineering fails to address issues in \bench, such as those involving dynamics.

\bibliography{iclr2025_conference}

\begin{thebibliography}{40}
\providecommand{\natexlab}[1]{#1}
\providecommand{\url}[1]{\texttt{#1}}
\expandafter\ifx\csname urlstyle\endcsname\relax
  \providecommand{\doi}[1]{doi: #1}\else
  \providecommand{\doi}{doi: \begingroup \urlstyle{rm}\Url}\fi

\bibitem[Pik(2023)]{Pika}
Pika, 2023.
\newblock URL \url{https://www.pika.art/}.

\bibitem[gen(2024)]{gen3}
Gen-3, 2024.
\newblock URL \url{https://runwayml.com/blog/introducing-gen-3-alpha/}.

\bibitem[kli(2024)]{kling}
Kling, 2024.
\newblock URL \url{https://kling.kuaishou.com/}.

\bibitem[Achiam et~al.(2023)Achiam, Adler, Agarwal, Ahmad, Akkaya, Aleman, Almeida, Altenschmidt, Altman, Anadkat, et~al.]{achiam2023gpt}
Josh Achiam, Steven Adler, Sandhini Agarwal, Lama Ahmad, Ilge Akkaya, Florencia~Leoni Aleman, Diogo Almeida, Janko Altenschmidt, Sam Altman, Shyamal Anadkat, et~al.
\newblock Gpt-4 technical report.
\newblock \emph{arXiv preprint arXiv:2303.08774}, 2023.

\bibitem[Bansal et~al.(2024)Bansal, Lin, Xie, Zong, Yarom, Bitton, Jiang, Sun, Chang, and Grover]{bansal2024videophy}
Hritik Bansal, Zongyu Lin, Tianyi Xie, Zeshun Zong, Michal Yarom, Yonatan Bitton, Chenfanfu Jiang, Yizhou Sun, Kai-Wei Chang, and Aditya Grover.
\newblock Videophy: Evaluating physical commonsense for video generation.
\newblock \emph{arXiv preprint arXiv:2406.03520}, 2024.

\bibitem[Battaglia et~al.(2013)Battaglia, Hamrick, and Tenenbaum]{battaglia2013simulation}
Peter~W Battaglia, Jessica~B Hamrick, and Joshua~B Tenenbaum.
\newblock Simulation as an engine of physical scene understanding.
\newblock \emph{Proceedings of the National Academy of Sciences}, 110\penalty0 (45):\penalty0 18327--18332, 2013.

\bibitem[Brooks et~al.(2022)Brooks, Hellsten, Aittala, Wang, Aila, Lehtinen, Liu, Efros, and Karras]{brooks2022generating}
Tim Brooks, Janne Hellsten, Miika Aittala, Ting-Chun Wang, Timo Aila, Jaakko Lehtinen, Ming-Yu Liu, Alexei Efros, and Tero Karras.
\newblock Generating long videos of dynamic scenes.
\newblock \emph{Advances in Neural Information Processing Systems}, 35:\penalty0 31769--31781, 2022.

\bibitem[Fu et~al.(2024)Fu, He, Lu, Wang, and Roth]{fu2024commonsense}
Xingyu Fu, Muyu He, Yujie Lu, William~Yang Wang, and Dan Roth.
\newblock Commonsense-t2i challenge: Can text-to-image generation models understand commonsense?
\newblock \emph{arXiv preprint arXiv:2406.07546}, 2024.

\bibitem[Gao et~al.(2024)Gao, Yang, Chen, Chitta, Qiu, Geiger, Zhang, and Li]{gao2024vista}
Shenyuan Gao, Jiazhi Yang, Li~Chen, Kashyap Chitta, Yihang Qiu, Andreas Geiger, Jun Zhang, and Hongyang Li.
\newblock Vista: A generalizable driving world model with high fidelity and versatile controllability.
\newblock \emph{arXiv preprint arXiv:2405.17398}, 2024.

\bibitem[Halliday et~al.(2013)Halliday, Resnick, and Walker]{halliday2013fundamentals}
David Halliday, Robert Resnick, and Jearl Walker.
\newblock \emph{Fundamentals of physics}.
\newblock John Wiley \& Sons, 2013.

\bibitem[Harjono et~al.(2020)Harjono, Gunawan, Adawiyah, and Herayanti]{harjono2020interactive}
Ahmad Harjono, Gunawan Gunawan, Rabiatul Adawiyah, and Lovy Herayanti.
\newblock An interactive e-book for physics to improve students' conceptual mastery.
\newblock \emph{International Journal of Emerging Technologies in Learning (iJET)}, 15\penalty0 (5):\penalty0 40--49, 2020.

\bibitem[He et~al.(2024{\natexlab{a}})He, Xue, Liu, Lin, Gao, Lin, Qiao, Ouyang, and Liu]{he2024venhancer}
Jingwen He, Tianfan Xue, Dongyang Liu, Xinqi Lin, Peng Gao, Dahua Lin, Yu~Qiao, Wanli Ouyang, and Ziwei Liu.
\newblock Venhancer: Generative space-time enhancement for video generation.
\newblock \emph{arXiv preprint arXiv:2407.07667}, 2024{\natexlab{a}}.

\bibitem[He et~al.(2024{\natexlab{b}})He, Jiang, Zhang, Ku, Soni, Siu, Chen, Chandra, Jiang, Arulraj, et~al.]{he2024mantisscore}
Xuan He, Dongfu Jiang, Ge~Zhang, Max Ku, Achint Soni, Sherman Siu, Haonan Chen, Abhranil Chandra, Ziyan Jiang, Aaran Arulraj, et~al.
\newblock Mantisscore: Building automatic metrics to simulate fine-grained human feedback for video generation.
\newblock \emph{arXiv preprint arXiv:2406.15252}, 2024{\natexlab{b}}.

\bibitem[Hessel et~al.(2021)Hessel, Holtzman, Forbes, Bras, and Choi]{hessel2021clipscore}
Jack Hessel, Ari Holtzman, Maxwell Forbes, Ronan~Le Bras, and Yejin Choi.
\newblock Clipscore: A reference-free evaluation metric for image captioning.
\newblock \emph{arXiv preprint arXiv:2104.08718}, 2021.

\bibitem[Huang et~al.(2024)Huang, He, Yu, Zhang, Si, Jiang, Zhang, Wu, Jin, Chanpaisit, et~al.]{huang2024vbench}
Ziqi Huang, Yinan He, Jiashuo Yu, Fan Zhang, Chenyang Si, Yuming Jiang, Yuanhan Zhang, Tianxing Wu, Qingyang Jin, Nattapol Chanpaisit, et~al.
\newblock Vbench: Comprehensive benchmark suite for video generative models.
\newblock In \emph{Proceedings of the IEEE/CVF Conference on Computer Vision and Pattern Recognition}, pp.\  21807--21818, 2024.

\bibitem[Jassim et~al.(2023)Jassim, Holubar, Richter, Wolff, Ohmer, and Bruni]{jassim2023grasp}
Serwan Jassim, Mario Holubar, Annika Richter, Cornelius Wolff, Xenia Ohmer, and Elia Bruni.
\newblock Grasp: A novel benchmark for evaluating language grounding and situated physics understanding in multimodal language models.
\newblock \emph{arXiv preprint arXiv:2311.09048}, 2023.

\bibitem[Kaplan et~al.(2020)Kaplan, McCandlish, Henighan, Brown, Chess, Child, Gray, Radford, Wu, and Amodei]{kaplan2020scaling}
Jared Kaplan, Sam McCandlish, Tom Henighan, Tom~B Brown, Benjamin Chess, Rewon Child, Scott Gray, Alec Radford, Jeffrey Wu, and Dario Amodei.
\newblock Scaling laws for neural language models.
\newblock \emph{arXiv preprint arXiv:2001.08361}, 2020.

\bibitem[Kay et~al.(2017)Kay, Carreira, Simonyan, Zhang, Hillier, Vijayanarasimhan, Viola, Green, Back, Natsev, et~al.]{kay2017kinetics}
Will Kay, Joao Carreira, Karen Simonyan, Brian Zhang, Chloe Hillier, Sudheendra Vijayanarasimhan, Fabio Viola, Tim Green, Trevor Back, Paul Natsev, et~al.
\newblock The kinetics human action video dataset.
\newblock \emph{arXiv preprint arXiv:1705.06950}, 2017.

\bibitem[Li et~al.(2024)Li, Zhang, Zhang, Zhang, Li, Li, Ma, and Li]{li2024llava}
Feng Li, Renrui Zhang, Hao Zhang, Yuanhan Zhang, Bo~Li, Wei Li, Zejun Ma, and Chunyuan Li.
\newblock Llava-next-interleave: Tackling multi-image, video, and 3d in large multimodal models.
\newblock \emph{arXiv preprint arXiv:2407.07895}, 2024.

\bibitem[Liao et~al.(2024)Liao, Lu, Zhang, Wan, Wang, Zhao, Zuo, Ye, and Wang]{liao2024evaluation}
Mingxiang Liao, Hannan Lu, Xinyu Zhang, Fang Wan, Tianyu Wang, Yuzhong Zhao, Wangmeng Zuo, Qixiang Ye, and Jingdong Wang.
\newblock Evaluation of text-to-video generation models: A dynamics perspective.
\newblock \emph{arXiv preprint arXiv:2407.01094}, 2024.

\bibitem[Lin et~al.(2024)Lin, Pathak, Li, Li, Xia, Neubig, Zhang, and Ramanan]{lin2024evaluating}
Zhiqiu Lin, Deepak Pathak, Baiqi Li, Jiayao Li, Xide Xia, Graham Neubig, Pengchuan Zhang, and Deva Ramanan.
\newblock Evaluating text-to-visual generation with image-to-text generation.
\newblock \emph{arXiv preprint arXiv:2404.01291}, 2024.

\bibitem[Liu et~al.(2024{\natexlab{a}})Liu, Li, Wu, and Lee]{liu2024visual}
Haotian Liu, Chunyuan Li, Qingyang Wu, and Yong~Jae Lee.
\newblock Visual instruction tuning.
\newblock \emph{Advances in neural information processing systems}, 36, 2024{\natexlab{a}}.

\bibitem[Liu et~al.(2024{\natexlab{b}})Liu, Ren, Gupta, and Wang]{liu2024physgen}
Shaowei Liu, Zhongzheng Ren, Saurabh Gupta, and Shenlong Wang.
\newblock Physgen: Rigid-body physics-grounded image-to-video generation.
\newblock In \emph{European Conference on Computer Vision ECCV}, 2024{\natexlab{b}}.

\bibitem[Liu et~al.(2024{\natexlab{c}})Liu, Cun, Liu, Wang, Zhang, Chen, Liu, Zeng, Chan, and Shan]{liu2024evalcrafter}
Yaofang Liu, Xiaodong Cun, Xuebo Liu, Xintao Wang, Yong Zhang, Haoxin Chen, Yang Liu, Tieyong Zeng, Raymond Chan, and Ying Shan.
\newblock Evalcrafter: Benchmarking and evaluating large video generation models.
\newblock In \emph{Proceedings of the IEEE/CVF Conference on Computer Vision and Pattern Recognition}, pp.\  22139--22149, 2024{\natexlab{c}}.

\bibitem[Mazzaglia et~al.(2024)Mazzaglia, Verbelen, Dhoedt, Courville, and Rajeswar]{mazzaglia2024multimodal}
Pietro Mazzaglia, Tim Verbelen, Bart Dhoedt, Aaron Courville, and Sai Rajeswar.
\newblock Multimodal foundation world models for generalist embodied agents.
\newblock \emph{arXiv preprint arXiv:2406.18043}, 2024.

\bibitem[Reid et~al.(2024)Reid, Savinov, Teplyashin, Lepikhin, Lillicrap, Alayrac, Soricut, Lazaridou, Firat, Schrittwieser, et~al.]{reid2024gemini}
Machel Reid, Nikolay Savinov, Denis Teplyashin, Dmitry Lepikhin, Timothy Lillicrap, Jean-baptiste Alayrac, Radu Soricut, Angeliki Lazaridou, Orhan Firat, Julian Schrittwieser, et~al.
\newblock Gemini 1.5: Unlocking multimodal understanding across millions of tokens of context.
\newblock \emph{arXiv preprint arXiv:2403.05530}, 2024.

\bibitem[Salimans et~al.(2016)Salimans, Goodfellow, Zaremba, Cheung, Radford, and Chen]{salimans2016improved}
Tim Salimans, Ian Goodfellow, Wojciech Zaremba, Vicki Cheung, Alec Radford, and Xi~Chen.
\newblock Improved techniques for training gans.
\newblock \emph{Advances in neural information processing systems}, 29, 2016.

\bibitem[Soomro(2012)]{soomro2012ucf101}
K~Soomro.
\newblock Ucf101: A dataset of 101 human actions classes from videos in the wild.
\newblock \emph{arXiv preprint arXiv:1212.0402}, 2012.

\bibitem[Sun et~al.(2024)Sun, Huang, Liu, Wu, Xu, Li, and Liu]{sun2024t2v}
Kaiyue Sun, Kaiyi Huang, Xian Liu, Yue Wu, Zihan Xu, Zhenguo Li, and Xihui Liu.
\newblock T2v-compbench: A comprehensive benchmark for compositional text-to-video generation.
\newblock \emph{arXiv preprint arXiv:2407.14505}, 2024.

\bibitem[Swartz(1985)]{swartz1985concept}
Norman Swartz.
\newblock \emph{The concept of physical law}.
\newblock Cambridge University Press, 1985.

\bibitem[Tan et~al.(2024)Tan, Yang, Qin, and Li]{tan2024vidgen}
Zhiyu Tan, Xiaomeng Yang, Luozheng Qin, and Hao Li.
\newblock Vidgen-1m: A large-scale dataset for text-to-video generation.
\newblock \emph{arXiv preprint arXiv:2408.02629}, 2024.

\bibitem[Unterthiner et~al.(2018)Unterthiner, Van~Steenkiste, Kurach, Marinier, Michalski, and Gelly]{unterthiner2018towards}
Thomas Unterthiner, Sjoerd Van~Steenkiste, Karol Kurach, Raphael Marinier, Marcin Michalski, and Sylvain Gelly.
\newblock Towards accurate generative models of video: A new metric \& challenges.
\newblock \emph{arXiv preprint arXiv:1812.01717}, 2018.

\bibitem[Wang et~al.(2023)Wang, Chen, Ma, Zhou, Huang, Wang, Yang, He, Yu, Yang, et~al.]{wang2023lavie}
Yaohui Wang, Xinyuan Chen, Xin Ma, Shangchen Zhou, Ziqi Huang, Yi~Wang, Ceyuan Yang, Yinan He, Jiashuo Yu, Peiqing Yang, et~al.
\newblock Lavie: High-quality video generation with cascaded latent diffusion models.
\newblock \emph{arXiv preprint arXiv:2309.15103}, 2023.

\bibitem[Wang et~al.(2024)Wang, Li, Li, Yu, He, Chen, Pei, Zheng, Xu, Wang, et~al.]{wang2024internvideo2}
Yi~Wang, Kunchang Li, Xinhao Li, Jiashuo Yu, Yinan He, Guo Chen, Baoqi Pei, Rongkun Zheng, Jilan Xu, Zun Wang, et~al.
\newblock Internvideo2: Scaling video foundation models for multimodal video understanding.
\newblock \emph{arXiv preprint arXiv:2403.15377}, 2024.

\bibitem[Wood et~al.(2024)Wood, Ullman, Wood, Spelke, and Wood]{wood2024object}
Justin~N Wood, Tomer~D Ullman, Brian~W Wood, Elizabeth~S Spelke, and Samantha~MW Wood.
\newblock Object permanence in newborn chicks is robust against opposing evidence.
\newblock \emph{arXiv preprint arXiv:2402.14641}, 2024.

\bibitem[Xiang et~al.(2024)Xiang, Liu, Gu, Gao, Ning, Zha, Feng, Tao, Hao, Shi, et~al.]{xiang2024pandora}
Jiannan Xiang, Guangyi Liu, Yi~Gu, Qiyue Gao, Yuting Ning, Yuheng Zha, Zeyu Feng, Tianhua Tao, Shibo Hao, Yemin Shi, et~al.
\newblock Pandora: Towards general world model with natural language actions and video states.
\newblock \emph{arXiv preprint arXiv:2406.09455}, 2024.

\bibitem[Yang et~al.(2024)Yang, Teng, Zheng, Ding, Huang, Xu, Yang, Hong, Zhang, Feng, et~al.]{yang2024cogvideox}
Zhuoyi Yang, Jiayan Teng, Wendi Zheng, Ming Ding, Shiyu Huang, Jiazheng Xu, Yuanming Yang, Wenyi Hong, Xiaohan Zhang, Guanyu Feng, et~al.
\newblock Cogvideox: Text-to-video diffusion models with an expert transformer.
\newblock \emph{arXiv preprint arXiv:2408.06072}, 2024.

\bibitem[Yuan et~al.(2024)Yuan, Huang, Xu, Liu, Zhang, Shi, Zhu, Cheng, Luo, and Yuan]{yuan2024chronomagic}
Shenghai Yuan, Jinfa Huang, Yongqi Xu, Yaoyang Liu, Shaofeng Zhang, Yujun Shi, Ruijie Zhu, Xinhua Cheng, Jiebo Luo, and Li~Yuan.
\newblock Chronomagic-bench: A benchmark for metamorphic evaluation of text-to-time-lapse video generation.
\newblock \emph{arXiv preprint arXiv:2406.18522}, 2024.

\bibitem[Zheng et~al.(2024)Zheng, Peng, Yang, Shen, Li, Liu, Zhou, Li, and You]{opensora}
Zangwei Zheng, Xiangyu Peng, Tianji Yang, Chenhui Shen, Shenggui Li, Hongxin Liu, Yukun Zhou, Tianyi Li, and Yang You.
\newblock Open-sora: Democratizing efficient video production for all, March 2024.
\newblock URL \url{https://github.com/hpcaitech/Open-Sora}.

\bibitem[Zhu et~al.(2024)Zhu, Wang, Zhao, Min, Deng, Dou, Wang, Shi, Wang, Zhang, et~al.]{zhu2024sora}
Zheng Zhu, Xiaofeng Wang, Wangbo Zhao, Chen Min, Nianchen Deng, Min Dou, Yuqi Wang, Botian Shi, Kai Wang, Chi Zhang, et~al.
\newblock Is sora a world simulator? a comprehensive survey on general world models and beyond.
\newblock \emph{arXiv preprint arXiv:2405.03520}, 2024.

\end{thebibliography}
\bibliographystyle{iclr2025_conference}

\clearpage
\newpage
\appendix
% \section{Appendix}

\section{PhyGenBench Details} \label{appendix:physgenbench}

\subsection{Detailed Overview} 
% \newpage
% xinyu
\begin{wraptable}{r}{0.4\textwidth}
    %\captionsetup{justification=raggedright, singlelinecheck=false} 
    %\hfill 
    \centering
    \caption{Details of \bench}
    \label{tab:overview}
    \begin{tabular}{@{}lc@{}}
        \toprule
        \textbf{Statistic} & \textbf{Number} \\ \midrule
         Physical Laws & 27\\
        Domains & 4 \\
        \quad  Optics & 50 \\
        \quad Mechanics & 40 \\
        \quad Thermal & 30 \\  
        \quad Material Properties & 40 \\
        \midrule
         Total Captions & 160 \\
        Total T2V Models & 8 \\
        Total Generated Videos & 1280 \\
        \midrule
        Unique Objects & 165 \\
        Unique Actions & 42 \\
        Average Length of Caption & 18.75 \\ 
        \bottomrule
    \end{tabular}
    \end{wraptable}

A fine-grained analysis of the dataset is essential for a comprehensive understanding of the benchmark. As shown in Table \ref{tab:overview}, \bench covers $4$ major domains in physics, encompassing $27$ representative physical laws, which enables it to provide a more comprehensive and fine-grained evaluation of models' physical capabilities. 
We generated $1280$ videos by evaluating $8$ advanced models.
Additionally, our captions encompass totally $165$ unique objects and $42$ unique actions with an average length of $18.75$ words.

\subsection{Difference between Videophy and Ours}
VIDEOPHY~\cite{bansal2024videophy} comprises 688 curated simple prompts that focus on interactions between three types of physical materials: solid-solid, solid-fluid, and fluid-fluid, but lack annotations of physical laws. The dataset is designed to evaluate a model's understanding of physical commonsense, featuring a limited range of physical phenomenons such as rigid body interactions, fluid dynamics, and contact forces. We are better suited than Videophy for evaluating physical commonsense due to two significant differences.

\begin{figure*}[htbp]
  \centering
  \scalebox{0.99}{
  \includegraphics[width=\linewidth]{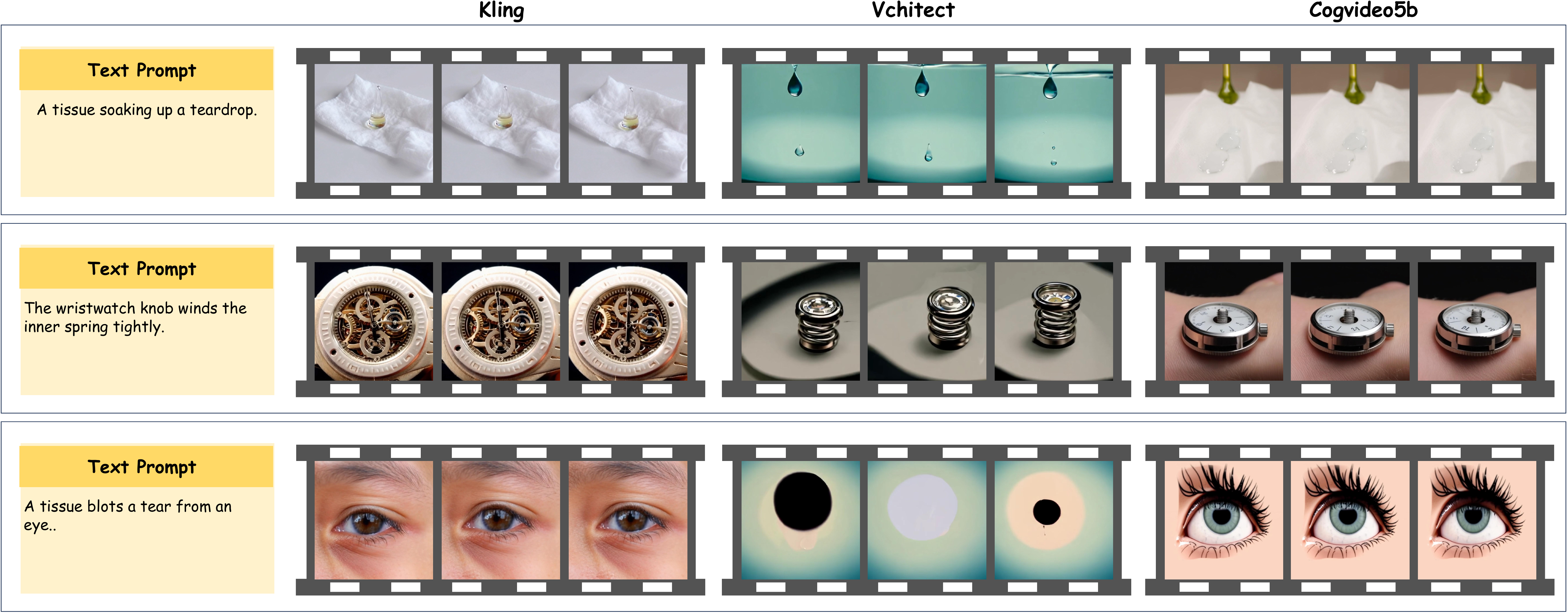}
  }
  \caption{\textbf{Samples of videos generated by Kling, Vchitect, and Cogvideo5b in Videophy.} All T2V models struggle to achieve proper text alignment and produce high-quality videos, making it meaningless to evaluate physical correctness in Videophy.}
  \label{fig:videophy}
  % \vspace{-10pt}
\end{figure*}

First As shown in Figure~\ref{fig:phybench}, \bench includes 160 carefully crafted prompts across 27 distinct physical laws, spanning four fundamental domains, which comprehensively assess a model’s understanding of physical commonsense. While Videophy primarily focuses on interactions between solid-fluid, solid-solid, and fluid-fluid, limiting its coverage and overlooking common physical laws such as phase transitions and basic material properties. What' more, Videophy lacks annotations of physical laws making it hard for VLM model to evaluate. Second, as shown in Table~\ref{tab:videophy-benchmark-ablation}, the average SA score of \bench ($0.80$) significantly outperforms that of Videophy ($0.63$). This indicates that \bench prompts are well-suited and easy for T2V models to generate high-quality, well-aligned videos, which benefits evaluation of physical correctness. In contrast, as shown in Figure~\ref{fig:videophy}, We find that prompts from Videophy pose challenges for T2V models in generating text-aligned and high-quality videos for two main reasons:  1. The prompts lack detail and specificity. For instance,\textit{``A tissue blots a tear from an eye"} is overly simplistic (without augmentation). Modern T2V models, such as CogVideo5B~\cite{yang2024cogvideox}, are typically trained with longer and more descriptive captions, which enhance their ability to comprehend and generate content based on prompts. 2. The scenes are often complex and unrealistic. For example, “The wristwatch knob winds the inner spring tightly” describes a process involving intricate internal mechanisms that are not visible externally. As a result, it is exceedingly difficult for T2V models to generate such scenes accurately.

\begin{table*}[h]
\centering
    \caption{\textbf{Comparison of SA results for video generation between Videophy and \bench.} We randomly select 64 prompts from both Videophy and \bench, use different T2V models to generate videos, and then ask annotators to score based on our cretiera in Figure \ref{fig:humaneval}. The results show that \bench’s SA scores significantly outperform Videophy.
} 
\label{tab:videophy-benchmark-ablation}
\resizebox{0.7\linewidth}{!}{%
\begin{tabular}%{lccc}
{lcccccccccccccccccc}
\toprule %\multirow{2}{*}
\multicolumn{1}{c}{{\textbf{Model}}} & \multicolumn{1}{c}{{\textbf{Size}}} & \multicolumn{1}{c}{\textbf{Videophy}($\uparrow$)}& \multicolumn{1}{c}{\bench($\uparrow$)}\\
\midrule
CogVideoX \citep{yang2024cogvideox} & 5B &$0.48$& $0.78$  \\
Vchitect 2.0 & 2B &$0.63$& $0.84$ \\ 
Kling & - & $\textbf{0.77}$ & $\textbf{0.89} $ \\
\midrule
Average & - & $0.63$ & $0.80 $ \\

\bottomrule
\end{tabular}
}
\end{table*}

% \subsection{construction}
% xinyu

\section{PhyGenEval Details} \label{appendix:PhysEval}
% Here we illustrate \eval with a more detailed version.

\subsection{Semantic alignment details} \label{appendix:SAdetail}
To reduce the complexity for VLM models to evaluate sementic correctness of generated videos between prompts, we adopt a two-stage strategy. Initially, we employ GPT-4o to extract objects and actions from the original text prompt. Subsequently, we employ GPT-4o to determine whether the extracted objects are present in the video and to verify the occurrence of specified actions. For each video, GPT-4o first assesses the presence of the objects mentioned in the prompt (e.g., \textit{``egg''}) within the video frames. This evaluation is performed according to Question 1 (Q1), where GPT-4o assigns a score from 0 to 2 based on the completeness of object presence: a score of 2 is given if all the objects are present, 1 if some of the objects are missing, and 0 if none of the objects appear in the video. After determining object presence, GPT-4o moves on to Question 2 (Q2) to check if the specified action (e.g., \textit{``pour out"}) is performed in the video. It assigns a binary score (0 or 1) depending on whether the action is present (1) or absent (0). Finally, these scores are combined to form the overall semantic alignment score. we put more details about other metric baselines in Appendix~\ref{appendix:sa-baseline}.

\subsection{Physical Commonsense alignment details}
In this section, we use the same notation as in Section \ref{phyevalpc} and provide a more detailed description of the calculation and design of the method.
% \textbf{Key Phenomena Detection. } Specifically, as shown in Figure \ref{fig:eval} (a), we first prompt GPT-4o to generate questions targeting key physical phenomena based on the input T2V prompt and corresponding physical laws, which include a retrieval prompt $P_r$ to locate the key frame and specific physical-based questions $Q$ about the key frame (if the ideal physical process is monotonic, the retrieval prompt defaults to the last frame). Next, we use CLIPScore \citep{hessel2021clipscore} to retrieve and locate the key frame from the video and employ VQAScore \citep{lin2024evaluating} to determine whether the expected physical phenomena are present in the key frame. Considering the semantic errors in the video, we also need to evaluate whether the retrieved key frame satisfies the retrieval prompt. Additionally, Considering only the top-1 retrieved frames is not robust enough (e.g., the egg may not be cracked just when it first contacts the stone). Therefore we consider using the $5$ frames around the key frame and jointly evaluating these frames.
\paragraph{Key Phenomena Detection. }  We categorize the T2V prompts into monotonic processes (eg. \textit{``melting with increasing temperature"}) and non-monotonic processes (eg. \textit{``an egg hitting a rock"}) based on the physical phenomena they represent. For prompt with monotonic processes, we only consider using the \textit{``Last Frame"} as the retrieval prompt, resulting in a single question. We can directly calculate $\mathrm{VLM}(\mathrm{Img}_j, Q)$, where the score for the corresponding video of this prompt ranges from 0 to 1. For prompt with  non-monotonic processes, we consider both the intermediate key frames and the Last Frame, resulting in two questions. For the intermediate key frames, we calculate $\mathrm{VLM}(\mathrm{Img}_j, Q) + \mathrm{VLM}(\mathrm{Img}_j, P_r)$, which ranges from 0-2. Consequently, the score range for videos corresponding to this prompt is 0 to 3.

For specific calculatation, we need to calculate $\mathrm{VLM}(\mathrm{I}_j, p_r)$ and $\mathrm{VLM}(\mathrm{I}_j, q)$, where $\mathrm{Img}_j$ is the $j$-th frame in the video. For $\mathrm{VLM}(\mathrm{I}_j, p_r)$, the calculation involves assessing the matching degree between the key frame and the retrieval prompt, which can be directly obtained using the original calculation method in \citep{lin2024evaluating}. For $\mathrm{VLM}(\mathrm{I}_j, q)$, we follow the computation approach from ChronoMagicBench \citep{yuan2024chronomagic}, we derive $\mathrm{VLM}(\mathrm{I}_j, q)$ by determining the ratio of the VQAScore for the affirmative statement to the combined VQAScores for both the affirmative and negative statements. We perform the calculations of $\mathrm{VLM}(\mathrm{I}_j, p_r)$ and $\mathrm{VLM}(\mathrm{I}_j, q)$ for each key frame within the specified range to obtain the physical correctness score for the problem.

\paragraph{Key Sequence Verification. } For this stage, which we've primarily introduced in Section \ref{sec:eval}, we focus on key calculation points. The score calculation formula for $q_1$ is $\mathrm{S_{before}} = \max_{i-2 \leq j \leq i} \left( \mathrm{VLM}(\mathrm{I}_0, \mathrm{I}_j, q_1) + \mathrm{VLM}(\mathrm{I}_j, p_r) \right)$. Here, $\mathrm{VLM}(\mathrm{I}_j, p_r)$ determines if the retrieved key frame satisfies the retrieval prompt,as the physical phenomenon should occur in the keyframe primarily located in Key Phenomena Detection, which is crucial for Key Sequence Verification (e.g.the expected physical phenomenon of egg cracking should occur in the keyframe when the egg hits the stone, rather than other frames when the egg is in the air or else). $\mathrm{VLM}(\mathrm{I}_0, \mathrm{I}_j, q_1)$ assesses the correctness of the Key Sequence order in the video. Notably, we calculate $\mathrm{VLM}(\mathrm{I}_j, p_r)$ using VQAScore, yielding a decimal between 0 and 1, while $\mathrm{VLM}(\mathrm{I}_0, \mathrm{I}_j, q_1)$ employs VLM (GPT-4V or LLaVA-Interleave) for question-answering, scoring 1 or 0 based on the model's Yes or No response.

\paragraph{Overall Naturalness Evaluation. } Here we mainly explain how to get the score of this part based on the evaluation results under the two-stage strategy described in Section \ref{sec:eval}. Specifically, we ask the video-based VLM to select the most appropriate option for the video according to the detailed scoring criteria generated by the LLM, and then we map the options to scores (Completely Fantastical to Almost Realistic corresponds to 0-3 points)

\paragraph{Overall Score. } We detail the discretization and calculation process of the scores here. In the stage of key phenomena detection, we categorize the prompts into monotonic and non-monotonic processes based on the physical phenomena they represent. For monotonic processes, the score range is 0-1, which we directly discretize by averaging into integer values from 0-3. Specifically, for non-monotonic processes with a score range of 0-3, we discretize the scores to $[1, 1.5, 2.25]$. This is because no points should be awarded if the physical phenomena are incorrect ($\mathrm{VLM}(\mathrm{I}_j, p_r)=1$ and $\mathrm{VLM}(\mathrm{I}_j, q)=0$), even with accurate retrieval. (e.g., The egg hits the stone and does not break)

In the stage of key sequence verification, we have three multi-image problems. One point is awarded for each correct answer, resulting in a final integer score from 0-3. Similar to the stage, of key phenomena detection we need to consider both the accuracy of key frame retrieval and the physical question answering. Therefore, we design the following: for $Q_1$, when $\max_{i-2 \leq j \leq i} \left( \mathrm{VLM}(\mathrm{I}_0, \mathrm{I}_j, q_1) + \mathrm{VLM}(\mathrm{I}_j, p_r) \right)$ and $\mathrm{VLM}(\mathrm{I}_j, p_r) > 0.5$, the question is considered correct. The process for $q_2$ is similar. For $q_3$, it is marked correct when $\mathrm{VLM}(\mathrm{I}_0, \mathrm{I}_{i-2: i+2}, \mathrm{I}_{-1}, q_3)$.

In the stage of overall naturalness evaluation, as we require video-based direct option selection, choosing Completely Fantastical, Clearly Unrealistic, Slightly Unrealistic, and Almost Realistic is scored as 0, 1, 2, and 3 points respectively. Finally, we average all scores and round down to obtain the final score.

% We use a four-point scale to rate physical correctness intuitively. For the single-image stage, if GPT raises one question with a retrieval prompt of "Last Frame" based on the prompt template, the score is directly averaged on a four-point scale. If two questions are raised, one retrieval prompt is "Last Frame" based on the prompt template, we put scores adjust to $[1, 1.5, 2.25]$ because no points should be awarded if the physical phenomena are incorrect ($VLM(img,re) = 1$, $VLM(img,qs) = 0$), even with accurate retrieval. For the multiple-image stage, each question scores $1$ point if VLM answers affirmatively and $VLM(img,re) > 0.5$, resulting in a four-level scale. For the video stage, the final score matches the VLM's choice on a four-level scale. We average scores from the final three stages, rounding to the nearest integer. Two scoring methods are used: \eval (GPT4o) using GPT-4o, and \eval (Open) using open-source models. The final score is their average. 

\section{Experiment} \label{appendix:experiment}

\subsection{Experiments Setup}

% \textbf{Human evaluation. } For human evaluation, we randomly select $64$ prompts from \bench and generate videos using $8$ T2V models, resulting in a total of $512$ videos requiring human assessment. For each video, we ask three annotators to provide both semantic and physical scores. We then average, round, and discretize the three scores for each video. Note that we ask the annotators to focus on the sementic alignment for sementic scores. Finally, we calculate the correlation between the human scores and automatic evaluation scores using Kendall’s $\tau$ and Spearman’s $\rho$.

\label{appendix:sa-baseline}
\paragraph{T2V model Implementation details. } Open-Sora 1.2~\citep{opensora} is an open-source project with the goal of reproducing Sora. CogVideoX 2b~\cite{yang2024cogvideox} and CogVideoX 5b are large-scale diffusion transformer models for text-to-video generation, incorporating a 3D Variational Autoencoder (VAE) for efficient video compression and an expert transformer with Expert Adaptive LayerNorm to improve text-video alignment. LaVie~\cite{wang2023lavie} is a cascaded video latent diffusion model. Vchitect2.0~\cite{wang2023lavie}, developed by the Shanghai AI Lab, is an advanced video generation model featuring a Parallel Transformer architecture to scale up video diffusion models and empower video creation.

\paragraph{Evaluation Metrics details. } We compare our proposed \eval with some evaluation metrics from previous methods like VideoPhy \citep{bansal2024videophy} and VideoScore \citep{he2024mantisscore}. VideoPhy fine-tunes a VLM with the VIDEOPHY dataset proposed by themselves, which includes human feed back about the semantic alignment and dynamic motion correctness about videos. VideoScore is trained on the VIDEOFEEDBACK dataset proposed by themselves, Initialized from the Mantis model. VideoScore provides automatic assessments of video quality based on human scoring criteria. To compare with \eval on SA and PCA, We only choose the text alignment and fact consistency criteria. Specifically, for the semantic alignment evaluation, we compare the Grid-LLaVA method proposed by T2V-CompBench, which extends the LLaVA \citep{liu2024visual} model to handle multi-frame inputs by sampling $6$ frames uniformly from a video to create an image grid.
For the physical commonsense alignment evaluation, we also compare with DEVIL \citep{liao2024evaluation}, which uses Gemini 1.5 Pro \citep{reid2024gemini} to assess the overall naturalness of videos and applies the same scoring standard prompt to all videos.

Furthermore, to evaluate the effectiveness of our \eval designs, we conduct a large amount of ablation studies and pue more details in Appendix \ref{appendix:abli}.

\begin{figure*}[t!]
  \centering
  \scalebox{0.8}{
  \includegraphics[width=\linewidth]{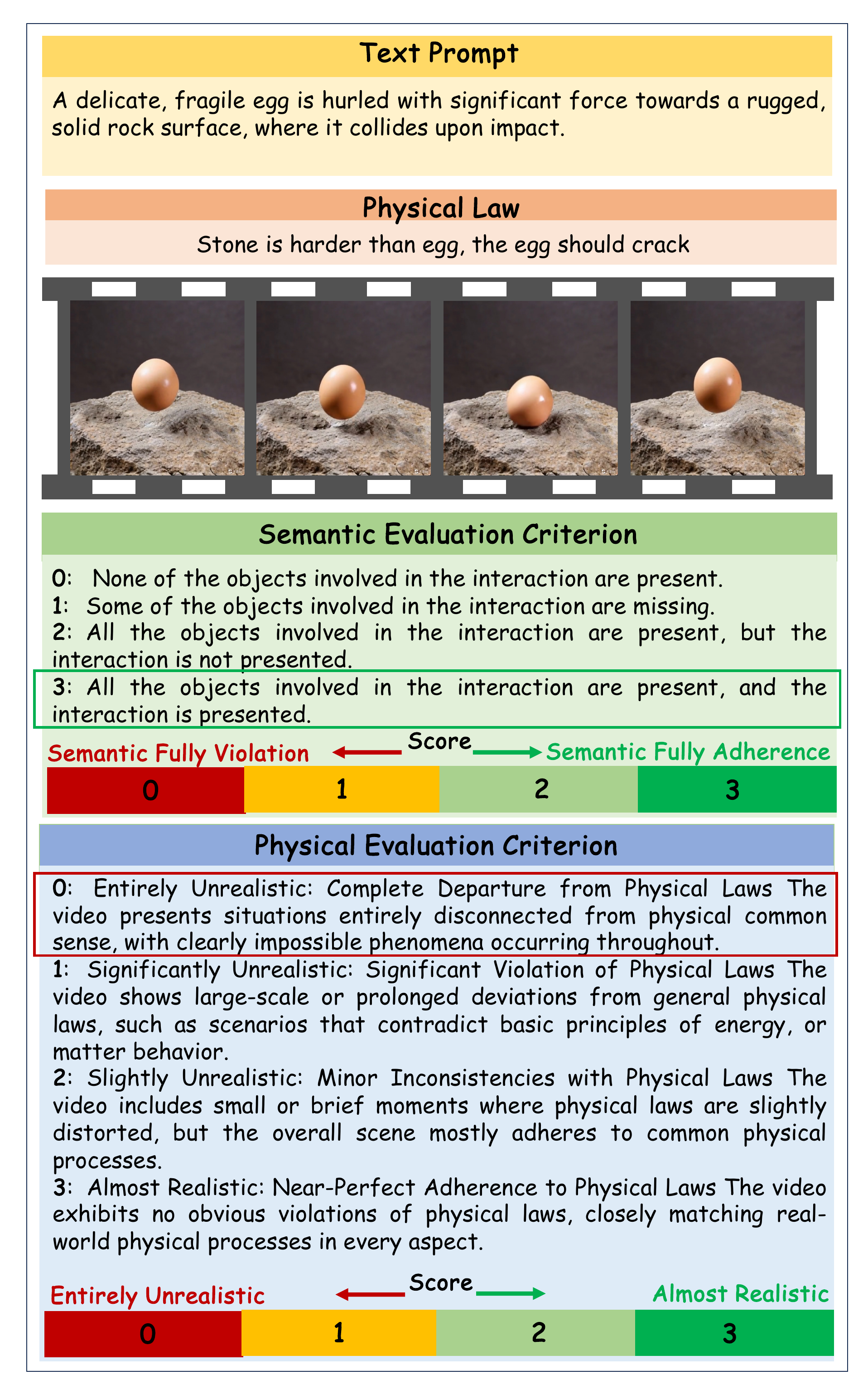}
  }
  \caption{Detailed diagram of the human evaluation process. We ask the annotators to score the semantic alignment and physical commonsense alignment of the video according to the scoring criteria in the figure.}
  \label{fig:humaneval}
  % \vspace{-10pt}
\end{figure*}

\paragraph{Human evaluation details. }\label{appendix:humaneval} Here, we provide a detailed explanation of the human evaluation described in Section \ref{main:humaneval}. Specifically, we require annotators to score based on the standards outlined in Figure \ref{fig:humaneval}, covering both semantic alignment and physical commonsense alignment. For example, as for the video shown in Figure \ref{fig:humaneval}, The egg bounces off the rock like a rubber ball, completely violating physical laws like dynamics, the annotator gives a score of $0$ for physical commonsense alignment. However, since the video fully includes the egg, the rock, and the collision action, the annotator gives a score of $3$ for semantic alignment.

\begin{table*}[t]
\vspace{-5pt}
\centering
    \caption{Details about evaluation models. The table shows duration, FPS, and resolution for each model.} 
\label{tab:fps}
\resizebox{0.8\linewidth}{!}{%
\begin{tabular}{lccc}
\toprule
\textbf{Model} & \textbf{Duration (s)} & \textbf{FPS} & \textbf{Resolution} \\
\midrule
Open-Sora 1.2 \citep{opensora} & 4 & 24 & 1280 $\times$ 720 \\
CogVideoX 2b & 6 & 8 & 720 $\times$ 480 \\
CogVideoX 5b & 6 & 8 & 640 $\times$ 360 \\
Lavie & 4 & 8 & 512 $\times$ 320 \\
Vchitect2.0 & 5 & 8 & 768 $\times$ 432 \\
\midrule
Pika \citep{Pika} & 3 & 24 & 1280 $\times$ 720 \\
Gen-3 \citep{gen3} & 11 & 24 & 1280 $\times$ 768 \\
Kling \citep{kling} & 5 & 30 & 1280 $\times$ 720 \\
\bottomrule
\end{tabular}
}
\vspace{-10pt}
\end{table*}

\subsection{Quantitative Evaluation} \label{appendix:quanti}

\paragraph{Comparison result about semantic alignment. }  Here we design a new baseline \eval(Grid-LLaVA) to illustrate the superiority of the method, which uses the two-stage strategy proposed in \eval from Appendix \ref{appendix:SAdetail}, but replaces the VLM with Grid-LLaVA proposed in T2V-CompBench \citep{sun2024t2v}. As shown in Table \ref{tab:SA_correlation}, \eval achieves the highest correlation scores across all categories, demonstrating its effectiveness as a human-aligned semantic commonsense correctness evaluator for \bench. Compared to other methods, \eval consistently outperforms previous baselines like VideoPhy, VideoScore, and Grid-LLaVA. Specifically, \eval obtains an overall Kendall's $\tau$ of $0.53$ and a Spearman's $\rho$ of $0.56$, surpassing the Grid-LLaVA ($\tau$: $0.35$, $\rho$: $0.39$). The results clearly show that our \eval design provides a more accurate and reliable semantic commonsense evaluation in \bench.

\begin{table}
\centering
    \caption{\textbf{SA correlation results with proposed \eval in video generation}. A higher score indicates better performance for a category. \textbf{Bold} stands for the best score, 
} 
\label{tab:SA_correlation}
\resizebox{\linewidth}{!}{%  
\begin{tabular}{l cccc cccc cc}
\toprule
\multirow{2}{*}{ \textbf{Metric}} & \multicolumn{2}{c}{\textbf{Mechanics}} & \multicolumn{2}{c}{\textbf{Optics}} & \multicolumn{2}{c}{\textbf{Thermal}} & \multicolumn{2}{c}{\textbf{Material}} & \multicolumn{2}{c}{\textbf{Overall}}\\
\cmidrule(lr){2-3}\cmidrule(lr){4-5}\cmidrule(lr){6-7}\cmidrule(lr){8-9}\cmidrule(lr){10-11}
 & $\tau$($\uparrow$) & $\rho$($\uparrow$) & $\tau$($\uparrow$) & $\rho$($\uparrow$) & $\tau$($\uparrow$) & $\rho$($\uparrow$)& $\tau$($\uparrow$) & $\rho$($\uparrow$) & $\tau$($\uparrow$) & $\rho$($\uparrow$)  \\
    \midrule
    VideoPhy~\citep{bansal2024videophy}& $0.20$ & $0.25$ & $0.03$& $0.03$& $0.20$& $0.24$& $0.18$& $0.22$& $0.13$& $0.17$\\
    VideoScore~\citep{he2024mantisscore}   & $0.14$ & $0.16$ & $-0.13$ & $-0.14$ & $0.23$ & $0.02$ & $0.02$& $0.02$& $0.05$& $0.05$\\
    Grid-LLaVA~\citep{sun2024t2v}   & $0.39$ & $0.43$ & $0.45$& $0.49$& $0.30$ & $0.33$ & $0.22$ & $0.26$ & $0.35$& $0.39$\\
    \eval(Grid-LLaVA)& $0.35$& $0.38$& $0.46$& $0.48$& $0.41$& $0.44$& $0.42$& $0.45$& $0.42$& $0.44$\\
    \eval& $0.48$ & $0.52$ & $0.64$ & $0.67$ & $0.46$ & $0.49$ & $0.47$ & $0.50$ & $0.53$ & $0.56$\\
    \bottomrule
\end{tabular}
}

\end{table}

\paragraph{Quantitative result about semantic alignment. }
As shown in Table \ref{tab:sa-benchmark} , nearly all models achieve relatively high SA scores, whether evaluated by machines or humans. This suggests that the scenarios in \bench are relatively straightforward, making it easier to assess physical commonsense. Among all the models, Kling achieved the highest SA score, with a human evaluation score of $0.89$, reflecting its strong instruction understanding and video generation capabilities. 

% Additionally, all models performed releatively better in the optics but worse in the material category. This indicates the need to gather more videos focused on material properties to improve the models' understanding of the physical characteristics of different materials.

\begin{table*}[t]
\centering
    \caption{\textbf{SA evaluation results with proposed \eval in video generation}. Both machine and human evaluations indicate that most models achieve good semantic scores on \bench. This suggests that the scenarios in \bench are simple enough to clearly reflect physical phenomena.
} 
\label{tab:sa-benchmark}
\resizebox{\linewidth}{!}{%
\begin{tabular}%{lccccccccccc}
{lcccccccccccccccccc}
\toprule %\multirow{2}{*}
\multicolumn{1}{c}{{\textbf{Model}}} & \multicolumn{1}{c}{{\textbf{Size}}} & \multicolumn{1}{c}{{\textbf{Mechanics}}($\uparrow$)} & \multicolumn{1}{c}{\textbf{Optics}($\uparrow$)} & \multicolumn{1}{c}{\textbf{Thermal}($\uparrow$)} & \multicolumn{1}{c}{\textbf{Material}($\uparrow$)} & \multicolumn{1}{c}{\textbf{Average}($\uparrow$)}  & \multicolumn{1}{c}{\textbf{Human}($\uparrow$)}\\
\midrule

CogVideoX \citep{yang2024cogvideox} & 2B  & $0.63$  & $0.67$  & $0.61$  &  $0.63$  &  $0.64$ & $0.64$  \\
CogVideoX \citep{yang2024cogvideox} & 5B  & $0.78$  & $0.88$  & $0.78$  &  $0.64$  &  $0.78$ & $0.78$  \\
Open-Sora V1.2 \citep{opensora} & 1.1B & $0.73$  & $0.85$  & $0.82$  &  $0.73$  &  $0.79$ & $0.70$ \\
Lavie \citep{wang2023lavie} & 860M & $0.47$  & $0.63$  & $0.73$  &  $0.53$  &  $0.58$ & $0.55$ \\ 
Vchitect 2.0 \citep{wang2023lavie} & 2B & $\textbf{0.92}$  & $0.89$  & $0.77$  &  $0.74$  &  $0.84$ & $0.84$ \\ 
\midrule
Pika \citep{Pika} & - & $0.63$  & $0.81$  & $0.73$  &  $0.69$  &  $0.72$ & $0.65$ \\
Gen-3 \citep{gen3} & - & $0.84$ & $\textbf{0.93}$  & $0.82$  &  $\textbf{0.78}$  &  $\textbf{0.85}$ & $0.86$ \\
Kling \citep{kling} & - & $0.88$  & $0.91$  & $\textbf{0.87}$ &  $0.74$  &  $\textbf{0.85}$ & $\textbf{0.89}$ \\

\bottomrule
\end{tabular}
}
\end{table*}

\subsection{Ablation study}\label{appendix:abli}

\paragraph{The Component in \eval on physical commonsense alignment evaluation. } We conduct a series of ablation studies to demonstrate the necessity of our method design by examining its correlation with human evaluation results, similar to those described in Section \ref{main:humaneval}. Specifically, we compare: 1) The effectiveness of two-stage evaluation method proposed in Section \ref{videoeval} 2) The effect of the various stages of \eval, as proposed in Section \ref{phyevalpc}; 3) Performance differences when using various VLMs and their ensembles in \eval, as outlined in Section \ref{phyevaloverall}. Notice that \eval for physical commnonsense alignment evaluation consists of three stages: key phenomena Detection, key sequence verification, and overall naturalness evaluation. And We denote them as \eval-S, \eval-M, and \eval-V based on the VLM they used.

1) We demonstrate that employing a two-stage strategy, as outlined in Section \ref{videoeval}, yields superior results when assessing the physical commonsense correctness of the entire video compared to one-stage strategy. 
Specifically, the one-stage strategy refers to not using LLM to rewrite the scoring template, but instead applying a single scoring template for all prompts' corresponding videos, allowing the VLM to score them. This method is proposed in DEVIL \citep{liao2024evaluation}. To verify the superiority of the two-stage strategy, we use InternVideo2 and GPT-4o as VLMs and perform both the one-stage and two-stage strategies. We label these as \eval-V(Intern) and \eval-V(GPT-4o), respectively. As shown in Table \ref{tab:twostage_abli}, the evaluation results produced by the two-stage strategy are more consistent with human judgments for both InternVideo2 and GPT-4o. We attribute this improvement to the incorporation of LLM (GPT-4o) for better comprehension of physical commonsense text, which effectively reduces the complexity of the task for VLMs in evaluating the physical correctness of videos.

2) \eval for physical commnonsense alignment evaluation consists of three stages. We investigate the contribution of each stage to the final performance. Table \ref{tab:eachstage_abli} presents results using one or two stages (employing ensemble strategies when multiple VLMs are applicable). We find that optimal performance is achieved only when all three stages are used concurrently, demonstrating the rationale behind \eval's design.

3) Given potential biases in single models and the costs associated with closed-source models, we offer two \eval computation methods: using GPT-4o or alternative open-source models (LLaVA-Interleave \citep{li2024llava} and InternVideo2 \citep{wang2024internvideo2}). Table \ref{tab:model_abli} shows that even using only open-source models achieves a high correlation coefficient of 0.66. Notably, ensembling both methods yields the best results. Considering \bench's relatively small size, we find this computational cost acceptable. Therefore we recommend users ensemble these methods.

\paragraph{The Component in \eval on semantic alignment evaluation. } we also perform necessary ablation experiments to validate the necessity of our SA evaluation design. Specifically, we compare: 1) VLM Model Selection: We leverage GPT-4o \citep{achiam2023gpt} as a more robust VLM model for SA evaluation. 2) Effectiveness of our two-stage evaluation method proposed in Appendix \ref{appendix:SAdetail}

1) As shown in Table \ref{tab:SA_correlation}, using GPT-4o in \eval is much better than using LLaVA, which achieve a higher Kendall's $\tau$ of $0.53$ compared to $0.42$, and a higher Spearman's $\rho$ of $0.56$ versus $0.44$. This indicates a stronger alignment between GPT-4o’s evaluations and human annotations  compared to open-source vlm models like Grid-LLaVA \citep{sun2024t2v}, justifying its selection as the preferred VLM model in the SA evaluation design. Since \bench includes a limited number of prompts, we believe that the cost of using GPT-4o is acceptable relative to the improvement in performance.

2) To validate the effectiveness of the two-stage strategy, we compare it with the method in T2V-CompBench \citep{sun2024t2v}, which directly uses Grid-LLaVA to apply the same scoring standard prompt for semantic alignment evaluation across all videos. For fairness, we also use Grid-LLaVA but implement the two-stage strategy proposed in Appendix \ref{appendix:SAdetail}. As shown in Table \ref{tab:SA_correlation}, \eval-Grid-LLaVA outperforms Grid-LLaVA, achieving a higher Kendall's $\tau$ score of 0.42 compared to 0.35, and a higher Spearman's $\rho$ score of 0.44 versus 0.39. This result demonstrates the effectiveness of our Two Stage Evaluation Method. By decomposing the evaluation into object detection and action detection, we effectively reduces the complexity of the task for VLMs in evaluating the sementic correctness of videos.

\begin{table}
\centering
    \caption{Comparison of PCA correlation results of the two-stage strategy for the video stage in \eval
} 
\resizebox{\linewidth}{!}{%  
\begin{tabular}{l cccc cccc cc}
\toprule
\multirow{2}{*}{ \textbf{Metric}} & \multicolumn{2}{c}{\textbf{Mechanics}} & \multicolumn{2}{c}{\textbf{Optics}} & \multicolumn{2}{c}{\textbf{Thermal}} & \multicolumn{2}{c}{\textbf{Material}} & \multicolumn{2}{c}{\textbf{Overall}}\\
\cmidrule(lr){2-3}\cmidrule(lr){4-5}\cmidrule(lr){6-7}\cmidrule(lr){8-9}\cmidrule(lr){10-11}
 & $\tau$($\uparrow$) & $\rho$($\uparrow$) & $\tau$($\uparrow$) & $\rho$($\uparrow$) & $\tau$($\uparrow$) & $\rho$($\uparrow$)& $\tau$($\uparrow$) & $\rho$($\uparrow$) & $\tau$($\uparrow$) & $\rho$($\uparrow$)  \\
    \midrule
    \multicolumn{11}{c}{\textbf{One Stage Strategy}} \\ 
    \midrule
    % one stage
    \eval-V(Intern) & $-0.03$ & $-0.04$ & $-0.20$ & $-0.21$ & $-0.26$ & $-0.27$ & $0.06$ & $0.06$ & $-0.10$ & $-0.11$ \\
    \eval-V(GPT)  & $0.39$ & $0.41$ & $0.11$ & $0.12$ & $0.19$ & $0.20$ & $0.36$ & $0.39$ & $0.19$ & $0.21$\\
    % \eval(Ensamble)  & $0.36$ & $0.39$ & $0.08$ & $0.09$ & $0.20$ & $0.20$ & $0.35$ & $0.38$ & $0.18$ & $0.19$\\
    \midrule
    \multicolumn{11}{c}{\textbf{Two Stage Strategy}} \\ 
    \midrule
    % two stage
    \eval-V(Intern)  & $0.01$ & $0.01$ & $0.06$ & $0.06$ & $0.08$ & $0.08$ & $0.10$ & $0.11$ & $0.07$ & $0.08$\\
    \eval-V(GPT)  & $0.47$ & $0.51$ & $0.50$ & $0.53$ & $0.46$ & $0.49$ & $0.53$ & $0.58$ & $0.53$ & $0.58$\\
    % \eval(Ensamble)  & $0.26$ & $0.30$ & $0.44$ & $0.47$ & $0.33$ & $0.35$ & $0.48$ & $0.52$ & $0.42$ & $0.46$\\

    \bottomrule
\end{tabular}
}
\label{tab:twostage_abli}
\end{table}

\begin{table}
\centering
    \caption{Comparison of PCA correlation results using each stage in \eval
} 
\resizebox{\linewidth}{!}{%  
\begin{tabular}{l cccc cccc cc}
\toprule
\multirow{2}{*}{ \textbf{Metric}} & \multicolumn{2}{c}{\textbf{Mechanics}} & \multicolumn{2}{c}{\textbf{Optics}} & \multicolumn{2}{c}{\textbf{Thermal}} & \multicolumn{2}{c}{\textbf{Material}} & \multicolumn{2}{c}{\textbf{Overall}}\\
\cmidrule(lr){2-3}\cmidrule(lr){4-5}\cmidrule(lr){6-7}\cmidrule(lr){8-9}\cmidrule(lr){10-11}
 & $\tau$($\uparrow$) & $\rho$($\uparrow$) & $\tau$($\uparrow$) & $\rho$($\uparrow$) & $\tau$($\uparrow$) & $\rho$($\uparrow$)& $\tau$($\uparrow$) & $\rho$($\uparrow$) & $\tau$($\uparrow$) & $\rho$($\uparrow$)  \\
    \midrule
    \eval-S  & $0.50$ & $0.54$ & $0.43$ & $0.45$ & $0.50$ & $0.54$ & $0.72$ & $0.77$ & $0.56$ & $0.61$\\
    \eval-M  & $0.46$ & $0.49$ & $0.49$ & $0.53$ & $0.55$ & $0.59$ & $0.53$ & $0.57$ & $0.55$ & $0.60$\\
    \eval-V  & $0.26$ & $0.30$ & $0.44$ & $0.47$ & $0.33$ & $0.35$ & $0.48$ & $0.52$ & $0.42$ & $0.46$\\
    % \eval-V & $0.47$ & $0.51$ & $0.50$ & $0.53$ & $0.46$ & $0.49$ & $0.53$ & $0.58$ & $0.53$ & $0.58$\\
    \eval-SM  & $0.58$ & $0.61$ & $0.47$ & $0.50$ & $0.58$ & $0.62$ & $0.66$ & $0.70$ & $0.60$ & $0.64$\\
    \eval-SV  & $0.56$ & $0.59$ & $0.41$ & $0.43$ & $0.58$ & $0.60$ & $0.70$ & $0.74$ & $0.59$ & $0.62$\\
    \eval-MV  & $0.50$ & $0.53$ & $0.50$ & $0.53$ & $0.53$ & $0.57$ & $0.60$ & $0.64$ & $0.57$ & $0.61$\\
    \eval  & \textbf{0.72} & \textbf{0.75} & \textbf{0.76} & \textbf{0.77} & \textbf{0.73} & \textbf{0.75} & \textbf{0.81} & \textbf{0.84} & \textbf{0.78} & \textbf{0.81} \\
    
    \bottomrule
\end{tabular}
}

\label{tab:eachstage_abli}
\end{table}

\begin{table}
\centering
    \caption{Comparison of PCA correlation results using different models such as GPT-4o or open-sourced models in \eval
} 
\resizebox{\linewidth}{!}{%  
\begin{tabular}{l cccc cccc cc}
\toprule
\multirow{2}{*}{ \textbf{Metric}} & \multicolumn{2}{c}{\textbf{Mechanics}} & \multicolumn{2}{c}{\textbf{Optics}} & \multicolumn{2}{c}{\textbf{Thermal}} & \multicolumn{2}{c}{\textbf{Material}} & \multicolumn{2}{c}{\textbf{Overall}}\\
\cmidrule(lr){2-3}\cmidrule(lr){4-5}\cmidrule(lr){6-7}\cmidrule(lr){8-9}\cmidrule(lr){10-11}
 & $\tau$($\uparrow$) & $\rho$($\uparrow$) & $\tau$($\uparrow$) & $\rho$($\uparrow$) & $\tau$($\uparrow$) & $\rho$($\uparrow$)& $\tau$($\uparrow$) & $\rho$($\uparrow$) & $\tau$($\uparrow$) & $\rho$($\uparrow$)  \\
    \midrule
    \eval(Open)  & $0.54$ & $0.57$ & $0.59$ & $0.62$ & $0.55$ & $0.58$ & $0.65$ & $0.69$ & $0.62$ & $0.66$\\
    \eval(GPT4o)  & $0.59$ & $0.63$ & $0.53$ & $0.57$ & $0.64$ & $0.68$ & $0.73$ & $0.77$ & $0.66$ & $0.71$\\
    \eval  & \textbf{0.72} & \textbf{0.75} & \textbf{0.76} & \textbf{0.77} & \textbf{0.73} & \textbf{0.75} & \textbf{0.81} & \textbf{0.84} & \textbf{0.78} & \textbf{0.81} \\

    \bottomrule
\end{tabular}
}
\label{tab:model_abli}
\end{table}

\section{Discussion} \label{appendix:discussion}

\begin{figure*}[htbp]
  \centering
  \scalebox{0.99}{
  \includegraphics[width=\linewidth]{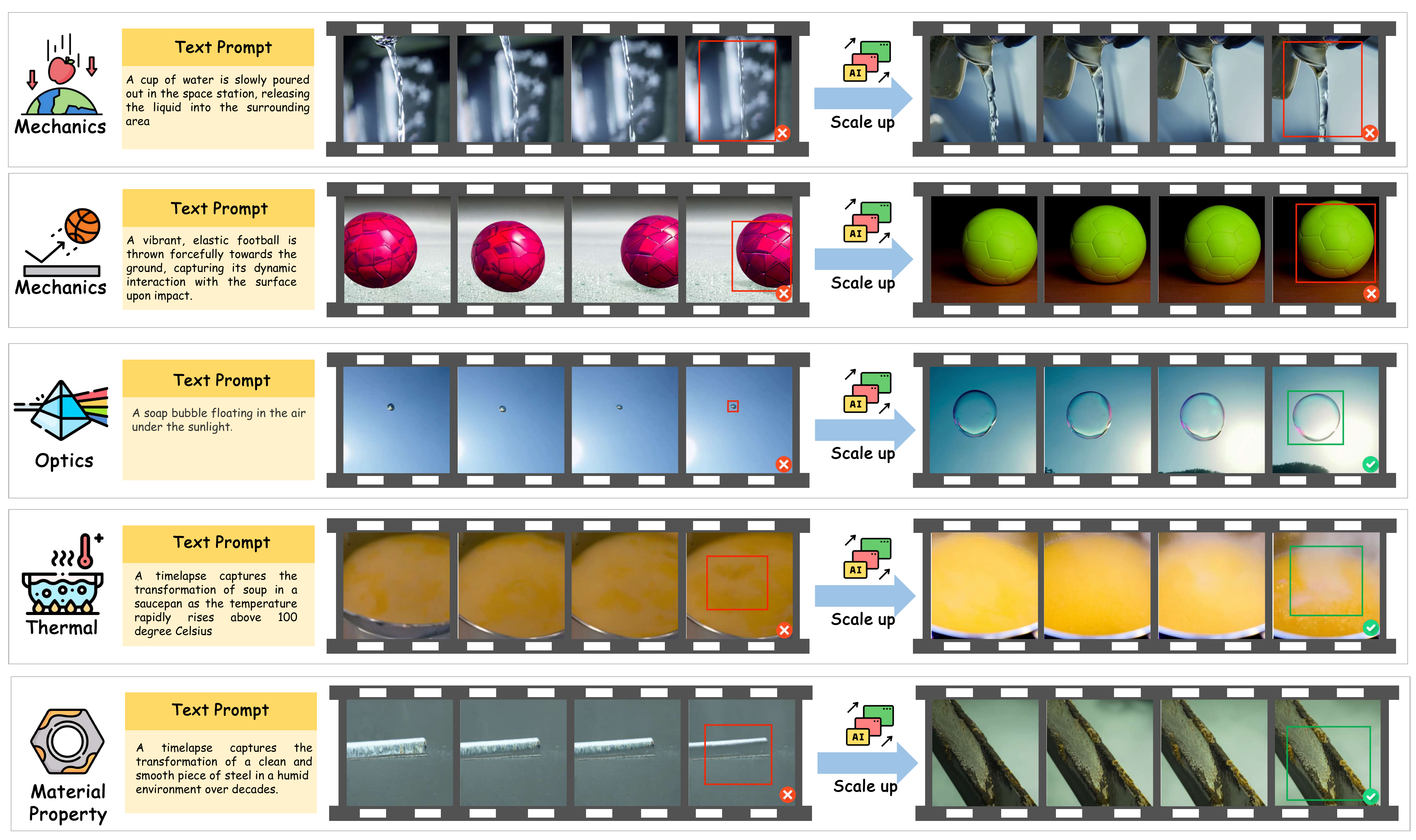}
  }
  \caption{The qualitative comparison of CogVideoX 2B and CogVideoX 5B. The result shows that simply scaling up can solve some issues, but dynamic physical phenomenons involving the design of motion patterns remain challenging.}
  \label{fig:scaleup}
  % \vspace{-10pt}
\end{figure*}

\paragraph{The Impact of Scaling on Physical Commonsense in Video Generation. } 
Scaling laws have been extensively validated in video generation models \citep{kaplan2020scaling}. We investigate their efficacy in addressing the challenges of physical commonsense presented in \bench. As shown in Table \ref{tab:pca-benchmark}, CogVideo 5B demonstrates improvements over CogVideo 2B, albeit with limited progress in the Mechanics category. Our qualitative analysis, illustrated in Figure \ref{fig:scaleup}, reveals significant advancements in static scenes with CogVideo 5B. It accurately captures complex phenomena such as colorful bubbles resulting from interference and diffraction, and oxidation-induced rusting of iron. In thermal, despite imperfections, CogVideo 5B generates more realistic boiling simulations compared to its predecessor. However, both models struggle with simple motion dynamics, exemplified by their inability to accurately depict a bouncing football. We posit that while scaling up enhances the model's capacity to generate videos that align with physical commonsense for individual objects, it may be insufficient for physical phenomenons involving dynamic physical laws. Addressing these challenges likely requires extensive training on carefully curated synthetic data, as suggested by \citep{liu2024physgen}. This approach could potentially bridge the gap in the model's grasp of fundamental physical laws.

\paragraph{Rewriting prompt. } We aim to explore whether GPT-augmented prompts can address the \bench challenges. Specifically, we rewrite the original prompts using GPT, adding expected physical outcomes and processes. For example, after \textit{``A bottle of juice is slowly poured out in the space station, releasing the liquid into the surrounding area"}, we add \textit{``The liquid forms floating globules, spreading out and moving randomly through the air."} in the end. 

\begin{figure*}[htbp]
  \centering
  \scalebox{0.99}{
  \includegraphics[width=\linewidth]{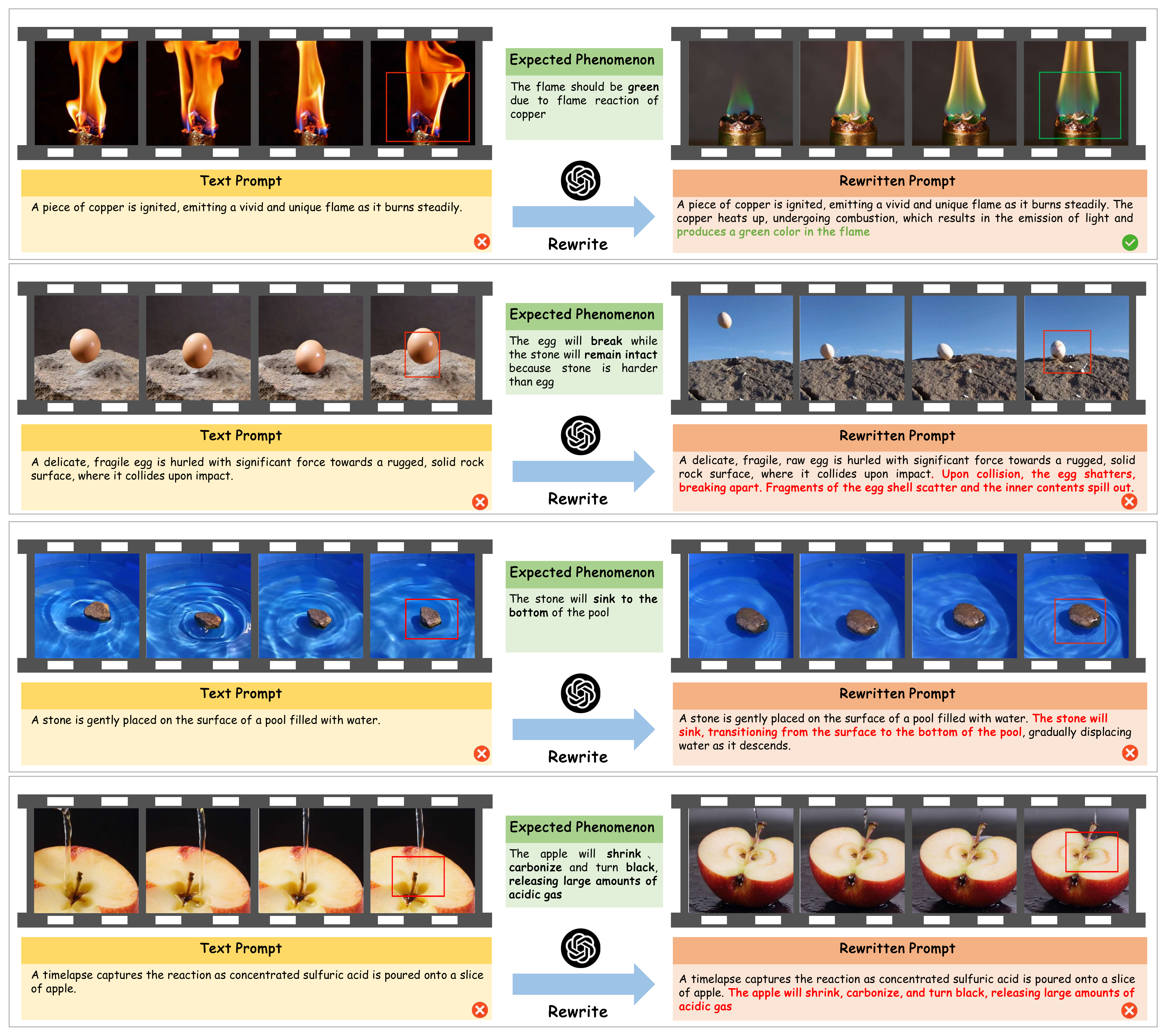}
  }
  \caption{The qualitative comparison of effects before and after using rewritten prompts. The results indicate that rewriting prompts addresses only a few basic issues (such as flame color reactions), while the majority of problems remain unsolved.}
  \label{fig:rewrite}
  \vspace{-10pt}
\end{figure*}

As shown in Table \ref{tab:rewrite}, we use CogVideoX 5b and Kling as representative models for open-source and closed-source systems, respectively, to conduct tests. The results indicate that prompt rewriting does help the models generate images aligned with physical laws, but it is still far from resolving the issues highlighted by \bench. Both CogVideoX 5b and Kling exhibit some growth, but even for Kling, it only achieves a score of 0.56. This demonstrates that current models still severely lack the ability to accurately render physical scenes, and this deficiency cannot be easily resolved through simple prompt rewriting. To illustrate this issue more clearly, as shown in Figure \ref{fig:rewrite}, our qualitative analysis shows that rewriting prompts can only address simple issues (e.g., flame color reactions), but remains ineffective for more complex physical processes (e.g., egg breaking, stone sinking).

\begin{table*}[t]
\centering
    \caption{\textbf{Evaluation results of PCA using the proposed \eval after rewriting prompts }. The results indicate that although using rewritten prompts leads to some improvement, it is still insufficient to address the challenges highlighted by \bench.
} 
\label{tab:rewrite}
\resizebox{\linewidth}{!}{%
\begin{tabular}%{lcccccccccc}
{lcccccccccccccccccc}
\toprule %\multirow{2}{*}
\multicolumn{1}{c}{{\textbf{Model}}} & \multicolumn{1}{c}{{\textbf{Size}}} & \multicolumn{1}{c}{{\textbf{Mechanics}}($\uparrow$)} & \multicolumn{1}{c}{\textbf{Optics}($\uparrow$)} & \multicolumn{1}{c}{\textbf{Thermal}($\uparrow$)} & \multicolumn{1}{c}{\textbf{Material}($\uparrow$)} & \multicolumn{1}{c}{\textbf{Average}($\uparrow$)}    \\
\midrule
\multicolumn{7}{c}{\textbf{Before Rewriting Prompt}} \\ 
\midrule
CogVideoX \citep{yang2024cogvideox} & 5B  & $0.39$  & $0.55$  & $0.40$  &  $0.42$  & $0.45$   \\
Kling & - & $0.45$  & $0.58$  & $0.50$  &  $0.40$  &  $0.49$ \\
\midrule
\multicolumn{7}{c}{\textbf{After Rewriting Prompt}} \\ 
\midrule
CogVideoX \citep{yang2024cogvideox} & 5B  & $0.39$  & $0.62$  & $0.53$  &  $0.52$  &  $0.52$  \\
Kling & - & $0.50$  & $0.64$  & $0.61$  &  $0.48$  &  $0.56$ \\

\bottomrule
\end{tabular}
}
\end{table*}

\paragraph{The robustness of \bench and \eval. } VEnhancer \citep{he2024venhancer} is a generative space-time enhancement framework that improves existing videos by adding spatial details and synthetic motion in the temporal domain. After enhancement by VEnhancer, Vchitect2.0 shows significant improvement on VBench, even surpassing Kling. However, VEnhancer only enhances the visual quality of videos (e.g., making them more coherent and clear) without addressing the model's poor understanding of physical commonsense.

As shown in Table \ref{tab:venhancer}, Vchitect enhanced by VEnhancer still scores similarly to the original version on \bench. We calculate a high Spearman coefficient of $0.86$ between model scores on \bench before and after VEnhancer enhancement. This indicates that \eval primarily focuses on physical correctness and is robust to other factors affecting visual quality. Furthermore, it demonstrates that even if a model can generate videos with better general quality (e.g., ranking higher on VBench), it doesn't necessarily imply a better understanding of physical common sense. This highlights the distinction between \bench and benchmarks like VBench that evaluate video quality.

\begin{table*}[t]
\centering
    \caption{\textbf{PCA evaluation results with proposed \eval in videos after VEnhancer}. The results indicate that employing VEnhancer fails to enhance the model’s comprehension of physical commonsense.
} 
\label{tab:venhancer}
\resizebox{\linewidth}{!}{%
\begin{tabular}%{lcccccccccc}
{lcccccccccccccccccc}
\toprule %\multirow{2}{*}
\multicolumn{1}{c}{{\textbf{Model}}} & \multicolumn{1}{c}{{\textbf{Size}}} & \multicolumn{1}{c}{{\textbf{Mechanics}}($\uparrow$)} & \multicolumn{1}{c}{\textbf{Optics}($\uparrow$)} & \multicolumn{1}{c}{\textbf{Thermal}($\uparrow$)} & \multicolumn{1}{c}{\textbf{Material}($\uparrow$)} & \multicolumn{1}{c}{\textbf{Average}($\uparrow$)}    \\
\midrule
Vchitect 2.0 & 2B & $0.41$  & $0.56$  & $0.44$  &  $0.37$  &  $0.45$ \\ 
Vchitect 2.0 (Venhancer) & 2B & $0.41$  & $0.56$  & $0.42$  &  $0.38$  &  $0.45$ \\

\bottomrule
\end{tabular}
}
\end{table*}

\end{document}